\documentclass{article}

\usepackage[preprint, nonatbib]{neurips_2021}

\usepackage[utf8]{inputenc} % allow utf-8 input
\usepackage[T1]{fontenc}    % use 8-bit T1 fonts
\usepackage{hyperref}       % hyperlinks
\usepackage{url}            % simple URL typesetting
\usepackage{booktabs}       % professional-quality tables
\usepackage{amsfonts}       % blackboard math symbols
\usepackage{amsmath,amsfonts,amsthm,amssymb,float,graphicx,subfigure,listings}
\usepackage{nicefrac}       % compact symbols for 1/2, etc.
\usepackage{microtype}      % microtypography
\usepackage{xcolor}         % colors
\usepackage{algorithm, algorithmic}
\usepackage{wrapfig}

\newcommand{\Mcal}[1]{\mathcal{#1}}
\newcommand{\mbf}[1]{\mathbf{#1}}
\newcommand{\bsyb}[1]{\boldsymbol{#1}}

\title{On the Power of Multitask Representation Learning in Linear MDP}

\author{%
  Rui Lu \\
  Institute for Interdisciplinary Information Science \\
  Tsinghua University\\
  \texttt{lur16@mails.tsinghua.edu.cn} \\
  \And
  Gao Huang\\
  Department of Automation \\
  Tsinghua University \\
  \texttt{gaohuang@tsinghua.edu.cn}
  \And 
  Simon S. Du\\
  Paul G. Allen School of Computer Science and Engineering\\
  University of Washington\\
  \texttt{ssdu@cs.washington.com} \\
}

\begin{document}
\maketitle
\begin{abstract}
While multitask representation learning  has become a popular approach in reinforcement learning (RL), theoretical understanding of why and when it works remains limited. This paper presents analyses for the statistical benefit of multitask representation learning in linear Markov Decision Process (MDP) under a generative model. In this paper, we consider an agent to learn a representation function $\phi$ out of a function class $\Phi$ from $T$ source tasks with $N$ data per task, and then use the learned $\hat{\phi}$ to reduce the required number of sample for a new task. We first discover a \emph{Least-Activated-Feature-Abundance} (LAFA) criterion, denoted as $\kappa$, with which we prove that a straightforward least-square algorithm learns a policy which is $\tilde{O}\left(H^2\sqrt{\frac{\mathcal{C}(\Phi)^2 \kappa d}{NT}+\frac{\kappa d}{n}}\right)$ sub-optimal. Here $H$ is the planning horizon, $\mathcal{C}\left(\Phi\right)$ is $\Phi$'s complexity measure, $d$ is the dimension of the representation (usually $d\ll \mathcal{C}(\Phi)$) and $n$ is the number of samples for the new task. In multitask RL, $N T$ is usually sufficiently large and the bound is dominated by the second term.
Thus the required $n$ is $O(\kappa d H^4)$ for the sub-optimality to be close to zero, which is much smaller than $O(\mathcal{C}(\Phi)^2\kappa d H^4)$ in the setting without multitask representation learning, whose sub-optimality gap is $\tilde{O}\left(H^2\sqrt{\frac{\kappa \mathcal{C}(\Phi)^2d}{n}}\right)$. This theoretically explains the power of multitask representation learning in reducing sample complexity. Further, we note that to ensure high sample efficiency, the LAFA criterion $\kappa$ should be small. In fact, $\kappa$ varies widely in magnitude depending on the different sampling distribution for new task. This indicates adaptive sampling technique is important to make $\kappa$ solely depend on $d$. Finally, we provide empirical results of a noisy grid-world environment to corroborate our theoretical findings.
\end{abstract}

\section{Introduction}
%Representation Learning
Due to huge size of state space or action space, large sample complexity is a major problem for reinforcement learning (RL). In order to learn a good policy with less samples, multitask representation learning tries to find a joint low-dimensional embedding (feature extractor) from different but related tasks, and then uses a simple function (e.g. linear) on top of the embedding \cite{baxter2000model,caruana1997multitask,li2010contextual}.
The underlying mechanism is that since the tasks are related, we can extract the shared structure knowledge more efficiently from all these tasks than treating each task independently, and then utilize this representation function for new tasks.

%Sequential decision-making, examples
Empirically, representation learning has become a popular approach for improving sample efficiency across various machine learning tasks~\cite{bengio2013representation}.
In particular,representation learning has become increasingly more popular in reinforcement learning \cite{teh2017distral,taylor2009transfer,lazaric2011transfer,rusu2015policy,liu2016decoding,parisotto2015actor,higgins2017darla,hessel2019multi,arora2020provable,d'eramo2020sharing}.
% For example, many sequential decision-making tasks share the same environment but have different reward functions.
Thus a natural approach is to learn a succinct representation that describes the environment and then make decisions for different tasks on top of the learned representation.

%theory in representation learning, ask the question
While multitask representation learning is already widely applied in sequential RL problems empirically, its theoretical foundation is still limited.
This paper tackles a fundamental problem:
\begin{center}
\textbf{When does multitask representation learning \emph{provably} improve the sample efficiency of RL?}
\end{center}

\subsection{Our Contributions}
In this paper, we study one popular model, linear MDP~\cite{jin2019provably} with a generative model.
We theoretically and empirically study how multitask representation learning improves the sample efficiency.
\begin{itemize}
	\item We first present a straightforward least-square based algorithm to learn a linear representation for multitask linear MDP, and then use this representation function to learn Q-value for target task. Theoretically, we show it reduces sample efficiency to $O(d^2)$ compared with $O(D^2d^2)$ for standard approach without using representation learning, where $D$ is the ambient input dimension and $d \ll D$ is the representation dimension.
	To our knowledge, this is the first result showing representation learning provably improves the sample efficiency in RL. We also generalize representation function to general non-linear cases.
	\item Our result also highlights the importance of a criterion on the sampling distribution $\mathcal{D}$ applied to the target task, which we called \emph{Least-Activated-Feature-Abundance}. Define $\kappa_{\Mcal{D}}=\lambda_{\min}^{-1}(\mathbb{E}_{x\sim \Mcal{D}}[\hat{\phi}(x)\hat{\phi}(x)^{\top}])$ where $\lambda_{\min}$ means the smallest eigenvalue of a matrix and $\hat{\phi}$ is learned representation function. We find that needed sample for new task is proportional to $\kappa_{\Mcal{D}}$. This indicates adaptive sampling may become necessary in RL to boost sample efficiency, otherwise when $\kappa_{\Mcal{D}}$ is large, the required sample complexity for new task is still large even if representation $\hat{\phi}$ is almost perfect. This is absent in supervised learning. 
	\item  We conduct experiments on a noisy grid-world domain, where the observation contains much redundant information, and empirically verify our theoretical findings.
\end{itemize}

\section{Related Work}
In this section, we review related theoretical results. 

%supervised learning
In the  supervised learning setting, there is a march of works on multitask learning and representation learning with various assumptions \cite{baxter2000model,du2017hypothesis,ando2005framework,ben2003exploiting,maurer2006bounds,cavallanti2010linear,maurer2016benefit,du2020few,tripuraneni2020provable}.
All these results assumed the existence of a common representation shared among all tasks.
However, this assumption alone is not sufficient.
For example, Mauer et al. \cite{maurer2016benefit} further assumed every task is i.i.d. drawn from an underlying distribution.
Recently, Du et al. \cite{du2020few} replaced the i.i.d. assumption with a deterministic assumption on the input distribution.
Finally, it is worth mentioning that Tripuraneni et al. \cite{tripuraneni2020provable} gave the method-of-moments estimator and built the confidence ball for the feature extractor, which inspired our algorithm for the infinite-action setting.

% in contrast to previous work that focused on building the confidence ball for the losses on new tasks. 
%\simon{We didn't cite Chi Jin's paper? Jiaqi: cited now}

%\simon{multitask linear regression}
%\simon{representation learning}
%
%%
%On the theoretical side, \cite{baxter2000model} performed the first theoretical analysis and gave sample complexity bounds using covering numbers.
%%Our paper is inspired by the analysis by \cite{maurer2016benefit}.
%\cite{maurer2016benefit} and follow-up work gave analyses on the benefit of representation learning for reducing the sample complexity of the target task.
%Besides assuming a common representation for the source and target tasks, they assumed every task is i.i.d. drawn from an underlying distribution and can obtain an $O(\frac{1}{\sqrt{n_1}} + \frac{1}{\sqrt{T}})$ rate.
%As pointed out in \cite{maurer2016benefit}, the $\frac{1}{\sqrt{T}}$ dependence is not improvable even if $n_1 \rightarrow \infty$ because $\frac{1}{\sqrt{T}}$ is the rate of concentration for the distribution over tasks. 
%Our paper uses natural deterministic conditions on the linear predictors and can obtain an $O(\frac{1}{n_1T})$ fast rate, which can fully utilize all source data.
%Our work use some of the analysis techniques in these two work.

The benefit of representation learning has been studied in sequential decision-making problems, especially in RL domains.
Arora et al. \cite{arora2020provable} proved that representation learning can reduce the sample complexity of imitation learning.
D'eramo et al. \cite{d'eramo2020sharing} showed that representation learning can improve the convergence rate of value iteration algorithm.
Both works require a probabilistic assumption similar to that in Maurer et al. \cite{maurer2016benefit} and the statistical rates are of similar forms as those in \cite{maurer2016benefit}.
Recently, Yang et al. \cite{yang2021impact} showed multitask representation learning reduces the regret in linear bandits, using the framework developed by Du et al. \cite{du2020few}.

We remark that representation learning is also closely connected to meta-learning \cite{schaul2010metalearning}, which also has a line of work of its theoretical properties. \cite{denevi2019learning,finn2019online,khodak2019adaptive,lee2019meta,bertinetto2018meta}.
%\cite{raghu2019rapid} empirically suggested that the effectiveness of meta-learning is due to its ability to learn a useful representation.
We also note that there are analyses for other representation learning schemes \cite{arora2019theoretical,mcnamara2017risk,galanti2016theoretical,alquier2016regret,denevi2018incremental}, which are beyond the scope of this paper.

%linear MDP
Linear MDP~\cite{yang2019sample,jin2019provably} is a popular model in RL, which uses linear function approximation to generalize large state-action space.
This model assumes both the transition and the reward is a linear function of given features.
Provably efficient algorithms have been provided in both the generative model setting and the online setting.
In this paper we study a natural multitask generalization of this model.

\section{Problem Setup }
\subsection{Notations}
Let $[n]=\{1,2,\hdots,N\}$. Use $\|\cdot\|$ or $\|\cdot\|_2$ to denote the $\ell_2$ norm for a vector or the spectral norm for a matrix. Denote $\|\cdot\|_F$ as the Frobenius norm of a matrix. Let $\langle\cdot,\cdot\rangle$ be the Euclidean inner product between vectors or matrices. Denote $I$ as identity matrix. For a matrix $A\in\mathbb{R}^{m\times n}$, let $\sigma_i(A)$ be its $i$-th largest singular value. Let $\mathrm{span}(A)$ be the subspace of $\mathbb{R}^m$ spanned by the columns of $A$, namely $\mathrm{span}(A)=\{Av:v\in\mathbb{R}^n\}$. Denote $P_A=A(A^{\top}A)^{\dagger}A^{\top}\in \mathbb{R}^{m\times m}$, which is the projection matrix onto $\mathrm{span}(A)$. Here $\dagger$ means the Moore-Penrose pseudo-inverse.  Also define $P_A^{\perp}=I-P_A$, which is the projection matrix onto $\mathrm{span}(A)^{\perp}=\{v|v^{\top}A=0,v\in\mathbb{R}^m\}$. For a positive semi-definite matrix B, which is denoted as $B\succeq 0$, denote its largest and smallest eigenvalue as $\lambda_{\max}(B)$ and $\lambda_{\min}(B)$. Let $B^{1/2}\succeq 0$ be the square root of $B$ which satisfies $(B^{1/2})^2=B$. We use the standard $O(\cdot),\Omega(\cdot)$ and $\Theta(\cdot)$ to denote the asymptotic bound ignoring the constant. Equivalently, we use notation $a\lesssim b$ to indicate $a=O(b)$ and use $a\ll b$ or $b\gg a$ to mean that $b\geq C\cdot a$ for a sufficient large universal constant $C>0$. We use $\Delta(S)$ to denote the set of all possible distribution supported on $S$.

A state space $\Mcal{S}$ is the set of all possible observations, and action space $\Mcal{A}$ is the set of all actions. A policy $\pi:\Mcal{S}\mapsto \Delta(\Mcal{A})$ is a mapping from state to action distribution. If the policy is not stationary, we use $\pi_h$ to denote the policy at level $h$. Transition probability $P_{(s,a)}$ is a probability measure supported on $\Mcal{S}$, which is the distribution of next state after performing action $a$ at state $s$ for a particular MDP. Reward function returns a value for any state-action pair $r(s,a):\Mcal{S}\times\Mcal{A}\mapsto \mathbb{R}$. Define Bellman operator $[\Mcal{T}_{a}g](s)=\mathbb{E}_{s'\sim P(s,a)}[g(s')]$ to simplify the notation. For any particular policy $\pi$, the value function $V_h^{\pi}(\cdot)$ at level $h$ is defined by $ V^{\pi}_h(s) = \mathbb{E}[ \sum_{k=h}^H r_{k}(s_h,\pi(s_h)) | s_h=s ]$, and corresponding Q-value function is defined by $Q_h^{\pi}(s,a)=r(s,a)+[\Mcal{T}_{a}V^{\pi}_{h+1}](s)$. The optimal policy which maximizes $V^{\pi}_h(s)$ for arbitrary $s\in\Mcal{S},h\in[H]$ is denoted as $\pi^{\star}$. Its corresponding value and Q-value function is denoted as $V^{\star}_h(\cdot)$ and $Q^{\pi^{\star}}_h(\cdot, \cdot)$.

\subsection{Linear MDP}
In this paper, we study a special class of MDP called \emph{linear MDP}. It permits a set of linear additive features $\phi_k(s,a),k=1,2,\hdots,d$ that fully describes the transition model $P_{s,a}$ and reward function $r(s,a)$. 
Formally, the learning agent is aware of $d$ feature functions $\phi_1,\phi_2,\hdots,\phi_d:\Mcal{S}\times\Mcal{A}\to \mathbb{R}$ or vector form $\phi:\Mcal{S}\times\Mcal{A}\to\mathbb{R}^d$ where
\begin{align}
    \phi(s,a)&=[\phi_1(s,a),\phi_2(s,a),\hdots,\phi_d(s,a)]\in \mathbb{R}^d.
\end{align}
These feature functions can be viewed as the structure knowledge for a specific finite-horizon MDP. Notice that in reality $\phi_k(x)$ takes input as a $D$-dimensional vector and we omit this agent-aware vectorization procedure $x=\xi(s,a)\in\mathbb{R}^D$. In all following analyses, we will directly write the feature mapping as $\phi(s,a)=\phi(\xi(s,a))$. 

We say an MDP is a linear MDP if the following assumption holds. 

\textbf{Linear MDP Assumption.} A finite-horizon MDP $\Mcal{M}=(\Mcal{S},\Mcal{A},H,P,r)$ is a linear MDP, if there exists a feature embedding function $\phi:\Mcal{S}\times\Mcal{A}\to \mathbb{R}^d$ and $\psi:\Mcal{S}\to \mathbb{R}^d,\theta\in\mathbb{R}^d$, such that the transition probability and reward function can be expressed as
\begin{align}
    P_{s,a}(\cdot)&=\langle\phi(s,a),\psi(\cdot)\rangle,\quad r(s,a)=\langle\phi(s,a),\theta\rangle.
\end{align}

Without loss of generality, we can normalize each $\phi(s,a)$ so that $\|\phi_k(s,a)\|\leq 1, \forall (s,a)\in\Mcal{S}\times\Mcal{A},k\in [d]$. Also, we can assume that $r(s,a)\leq 1$ by dividing $\max_{s,a}[r(s,a)]$ for all $r(s,a)$.

\textbf{Remark 1.} Linear MDP has an important property that the Q-value induced by any given policy $\pi$ can be simply calculated from a weight vector $\bsyb{w}^{\pi}_h$ as $Q^{\pi}_h(s,a)=\langle\bsyb{w}_h^{\pi},\phi(s,a)\rangle$. Since
\begin{align*}
     Q^{\pi}_{h}(s,a)&=r(s,a)+[\Mcal{T}_{\pi_h(s)} V^{\pi}_{h+1}](s)\\
     &=\phi(s,a)^{\top}\theta+\sum\nolimits_{s'\in\Mcal{S}}V^{\pi}_{h+1}(s')\cdot(\phi(s,a)^{
     \top}\psi(s'))\\
     &=\langle\phi(s,a),\theta+\sum\nolimits_{s'\in\Mcal{S}}V^{\pi}_{h+1}(s')\cdot\psi(s')\rangle.
\end{align*}
This means for linear MDP, given the feature extracting function $\phi(s,a)$, learning Q-value function is equivalent to learning a single weight vector $\bsyb{w}_h$ for each step $h\in[H]$. Choosing the action $a$ which maximizes $\langle\phi(s,\cdot),\bsyb{w}_h\rangle$ will give out the optimal policy action at step $h$ automatically.

\textbf{Remark 2.}\label{point:rmk2} Another important remark is that such property solely depends on the reward function $r(s,a)$ and transition dynamics $P(s,a)$, regardless of value function $V(\cdot)$. So for any value function $V_{h+1}(\cdot)$, the $Q$-value function can always be represented as $\tilde{Q}_h(s,a)=\langle\phi(s,a),\tilde{\bsyb{w}}_h\rangle$ for certain weight vector $\tilde{\bsyb{w}}_h$ without changing the representation $\phi(\cdot,\cdot)$.

\subsection{Multitask Learning Procedure}
In this section we will formally describe the problem setting that we aim to study.

Consider an agent performing actions in $T$ different tasks $\mu^{(t)},t\in[T]$ which are sampled i.i.d. from a multitask distribution $\eta$. Each task $\mu$ in the support set of $\eta$ is solving linear MDP $\Mcal{M}_{j}=(\Mcal{S},\Mcal{A},H,P_{j},r_{j})$. All these tasks $\mu$ share the same state space $\Mcal{S}$, action space $\Mcal{A}$, planning horizon $H$ and feature mapping $\phi(\cdot,\cdot)$. We hope to learn policies from function class $\Pi=\Mcal{F}\circ\Phi$, where $\Phi\subseteq \{\phi:\Mcal{S}\times\Mcal{A}\to\mathbb{R}^d|\quad \|\phi(s,a)\|_{\infty}\leq 1\}$ is a class of bounded norm \textit{representation functions} mapping from state-action pair to a latent representation, and $\Mcal{F}\subseteq\{f:\mathbb{R}^{d}\to\mathbb{R}\}$ is a mapping from representation to value estimation, i.e. in our setting this function class is simply $\Mcal{F}=\{x\to \bsyb{w}^{\top}x\}$. A policy $\pi^{\phi,f}$ is parametrized by representation $\phi\in\Phi$ and $f\in\Mcal{F}$, where $\pi^{\phi,f}(s)=\arg\max_{a\in\Mcal{A}}f(\phi(s,a))$. Our algorithm actually learns Q-value mapping  $Q^{\phi,f}(s,a)=f(\phi(s,a))$. 

The detail of our learning algorithm is shown in \hyperref[alg:train]{Alg 1} and \hyperref[alg:Test]{Alg 2}.

\begin{algorithm}[ht]
   \caption{Representation Learning for Training Tasks}
   \label{alg:train}
    \begin{algorithmic}[1] 
        \STATE $V^{(t)}_{H+1}(s)=0,\forall t\in[T],s\in\Mcal{S}$
        \STATE Sample $\mu^{(t)}\sim \eta,\quad t\in[T]$
        \FOR {$h:H\to 1$}
            \FOR{$t:1\to T$}
                \STATE Construct $\Mcal{D}^t$ for task $\mu^{(t)}$
                \STATE Sample $(s^t_j,a^t_j)\sim \Mcal{D}^{t}, \quad j\in[N]$
                \STATE $m\gets (H-h+1)^2$
                \STATE Query $(s_j^t,a_j^t)$ for $m$ times and get $r_{jk}\gets r^{(t)}(s^t_j,a^t_j)$,
                $\quad s'_{jk}\sim P^{\mu^{(t)}}_{s^t_j,a^t_j},\quad  j\in[N],k\in[m]$
                \STATE $Q_h^{(t)}(s^t_j,a^t_j)=\frac{1}{m}\sum_{k=1}^m r_{jk}+V^{(t)}_{h+1}(s'_{jk})$
            \ENDFOR
            \STATE $
                \phi^{(h)},\bsyb{w}_{t,h}=\arg\min_{\phi,\bsyb{w}_{t,h}} \quad \frac{1}{2 N T}\sum_{t=1}^T\sum_{j=1}^N\left[\phi(s_j^t,a_j^t)^{\top}\bsyb{w}_{t,h}-Q^{(t)}_h(s^t_j,a^t_j)\right]^2
                $
            \STATE $V^{(t)}_h(s)=\max_{a\in\Mcal{A}}\left\langle\phi^{(h)}(s,a),\bsyb{w}_{t,h}\right\rangle,\quad \forall s\in\Mcal{S},t\in[T]$
        \ENDFOR
        \end{algorithmic}
\end{algorithm}
\begin{algorithm}[ht]
    \caption{Applying $\hat{\phi}$ for New Task}
    \label{alg:Test}
    \begin{algorithmic}[1] 
        \STATE $V_{H+1}(s)=0,\forall s\in\Mcal{S}$
        \STATE $\hat{\phi}(\cdot,\cdot)\gets$ Algorithm 1
        \FOR{$h:H\to 1$}
            \STATE Construct $\Mcal{D}^{T+1}$ w.r.t. $\hat{\phi}_h$
            \STATE sample $(s^t_j,a^t_j)\sim \Mcal{D}^{T+1},\quad j\in[n]$ 
            \STATE $m\gets (H-h+1)^2$
            \STATE Query $(s_j^t,a_j^t)$ for $m$ times and get $r_{jk}\gets r^{(T+1)}(s^t_j,a^t_j), s'_{jk}\sim P^{\mu^{(T+1)}}_{s^t_j,a^t_j}, j\in[n],k\in[m]$
            \STATE $Q_h(s^t_j,a^t_j)=\frac{1}{m}\sum_{k=1}^m r_{jk}+V_{h+1}(s'_{jk})$
            \STATE $\bsyb{w}_{T+1,h}=\arg\min_{\bsyb{w}} \quad \frac{1}{2n}\sum_{j=1}^n\left[\hat{\phi}(s_j^t,a_j^t)^{\top}\bsyb{w}-Q_h(s^t_j,a^t_j)\right]^2$
            \STATE $V_h(s)=\max_{a\in\Mcal{A}}\left\langle\hat{\phi}(s,a),\bsyb{w}_{T+1,h}\right\rangle,\quad \forall s\in\Mcal{S}$
        \ENDFOR
        \STATE $\pi_h(s)=\arg\max_{a} \left\langle\hat{\phi}(s,a),\bsyb{w}_{T+1,h} \right\rangle$
    \end{algorithmic}
\end{algorithm}
At each level $h$, the algorithm samples $N$ state-action pairs $(s_j,a_j)$ from certain constructed distribution $\Mcal{D}^t$ (which will be clarified \hyperref[asmp:data]{later}) and then use each of them to query the generative model of task $\mu^{(t)}$ for $m$ times to simulate the MDP. After that, compute estimated Q-value label $Q_h(s,a)$ from the summation of sampled rewards $r_{j,k}$ and estimation for future cumulative reward predicted by $\phi^{(h+1)}$ and $\bsyb{w}_{t,h+1}$. The algorithm learns representation $\phi^{(h)}(\cdot, \cdot)$ and $\bsyb{w}_{t,h}$ for $t\in[T]$ by optimizing
\begin{align*}
    \min_{\phi, \bsyb{w}_{t,h}}& \frac{1}{2 N T} \sum_{t=1}^T \sum_{j=1}^N \left[ \phi(s_j^t, a_j^t)^{\top} \bsyb{w}_{t,h} - Q^{(t)}_h(s^t_j, a^t_j) \right]^2,\\
    &s.t.\quad \phi\in\Phi, \bsyb{w}_{t,h}\in\mathbb{R}^d.
\end{align*}
In this paper, we assume we can solve this optimization problem and analyze its solution.
We note that requiring a computational oracle to solve this empirical risk minimization problem is common in the literature~\cite{du2020few,tripuraneni2020theory}, so the analysis can be more focused on the merit of $\phi$ without dealing with concrete optimization procedure in detail.
In practice, one can apply gradient-based algorithms for this optimization.

% An immediate question is how to solve such optimization problem. In this paper, we will assume that by using some gradient-based methods or others, the algorithm can get a near-optimal solution to the problem above. Concrete solving procedure will not be discussed in detail, since how to solve $\phi$ and how good is the empirically best $\phi$ with respect to limited samples are orthogonal problems. \textbf{TODO: How to shuaiguo}

After obtaining the learned representation $\hat{\phi}$, we apply this representation on a new task. An essential part of our algorithm is that the state-action pair distribution $\Mcal{D}^{(T+1)}$ may need to be specially constructed with respect to learned $\hat{\phi}$. Only when the direction of the least activated feature is abundant enough, namely $\lambda_{\min}(\mathbb{E}_{x\sim \Mcal{D}^{(T+1)}}[\hat{\phi}(x)\hat{\phi}(x)^{\top}])=\Omega(1/d)$, it is guaranteed that $n(\ll N)$ samples suffice to learn a high-quality weight vector $\bsyb{w}_{T+1,h}$ for the new task. Otherwise it may still take $n=\Omega(D)$ samples even if $\hat{\phi}$ is almost perfect (see our experiments in section \hyperref[sec:sec7]{7.4}).

The final theoretical evaluation metric for our learned policy  $\pi_h(s)=\arg\max_{a} \langle\hat{\phi}(s,a),\bsyb{w}_{T+1,h} \rangle$ is to measure its sub-optimality compared with $\pi_h^{\star}(s)=\arg\max_{a} \langle\phi(s,a),\bsyb{w}^*_{T+1,h} \rangle$. Denote the expected value of a state $s$ given all the following steps using policy $\pi$ as $V^{\pi}(s)$, we aim to prove that
$
    \left\|V^{\pi}(s)-V^{\pi^{\star}}(s)\right\|\text{ is small, for } \forall\ s\in \Mcal{S}.
$

\section{Assumptions}
\textbf{Generative model assumption.}\label{asmp:generative} At line 8 for \hyperref[alg:train]{Alg1} and line 7 for \hyperref[alg:Test]{Alg2}, we assume that the learning agent has access to a particular generative model to query any state-action pair $(s,a)$, getting reward $r(s,a)$ and one sample from $P_{s,a}$ for any $(s,a)\in\Mcal{S}\times\Mcal{A}$. The accessibility to such generative model is a common assumption for theoretical analysis \cite{azar2012sample, agarwal2020model, cui2020plug}, and it is not unrealistic when the learning agent can use physics engine or interactive tool to simulate the environment.

\textbf{Data Assumption.}\label{asmp:data} We assume that we can sample $x$ from a uniform distribution over supporting set $S$ as all standard basis in $\mathbb{R}^D$, namely $\mathcal{D}^t=\operatorname{uniform}\{e_i\in\mathbb{R}^D:i\in[D]\}$. Each $e_i$ corresponds to certain valid state-action pair $e_i=\xi(s_i,a_i)$ and those $x_i$ spans the whole $\mathbb{R}^D$ space. Notice this also implies our sampling distribution is 1-Sub-Gaussian. Also, we assume that environment feedback reward $r_s(s,a)$ is generated from $r_s(s,a)=r(s,a)+z$ where $r(s,a)$ is the ground truth reward and $z$ is the noise sampled from $z\sim \Mcal{Z}$. We assume that $|z|\leq \sigma$, which means the noise is constant bounded. For deterministic model, $\sigma=0$. Furthermore, we make following assumptions.

% namely the tail of the distribution is no more than a certain Gaussian distribution. 
% Also, since we use bary-centric coordinate, we naturally have the bound for input as $\|x\| \leq \|x\|_1 = 1$. 
\textbf{Assumption 5.1} (Bounded Input) \textit{For any possible $(s,a)\in\mathcal{S}\times \mathcal{A}$, $\|\xi(s,a)\|\leq 1$.}

\textbf{Assumption 5.2} (Task Regularity and Diversity) \textit{The optimal weight vector for each task $t$ at level $h$ satisfies $\|\bsyb{w}^{*}_{t,h}\|=\Theta(H-h+1)$. Also, $\overline{W^{*}_h}=\frac{1}{H-h+1}[\bsyb{w}^*_{1,h},\bsyb{w}^*_{2,h},\hdots,\bsyb{w}^*_{T,h}] \in \mathbb{R}^{D\times T}$ satisfies $\sigma_{d}^2(\overline{W_h^*})\geq \Omega(T/d)$, or  $W_h^*=[\bsyb{w}^*_{1,h},\bsyb{w}^*_{2,h},\hdots,\bsyb{w}^*_{T,h}]$ satisfy $\sigma_{d}^2(W_h^*)\geq \Omega(T(H-h+1)^2/d)$.} 

Assumption 5.1 assumes that the norm of all possible state-action is constant bounded, which means the MDP contains no irregular outliers. Assumption 5.2 makes constraints for the tasks. First sentence assumes that the optimal weight vector $\|\bsyb{w}^{*}_{h}\|=\Theta(H-h+1)$. Since the maximum cumulative from level $h$ to end is $H-h+1$, such regularization constraint essentially means the weight matrix's singulars are of the same order. This also implies $\sum_{j=1}^d \sigma_j^2(\overline{W^*_h})=\|\overline{W^*_h}\|^2_F=\Theta(T)$. So the diversity constraint is equivalent to saying $\frac{\lambda_{\max}({W^*_h})}{\lambda_{\min}({W^*_h})}=O(1)$, which basically means that $\{\bsyb{w}_t^*\}_{t\in[T]}$ spreads evenly in the $\mathbb{R}^d$ space. 
This assumption has been used in \cite{du2020few} for developing the theory for representation learning.

\section{Linear Representation Function Analysis}
\label{sec:linear}
In this section, we consider the case where the representation is a linear mapping $\phi$ from the original input space $\mathbb{R}^D$ to a low dimensional space $\mathbb{R}^d(d\ll D)$. More specifically, we constrain the feature mapping class as linear functions $\Phi=\{\phi(x)= B^{\top}x|B\in\mathbb{R}^{D\times d}:B^{\top}B=k\cdot I\}$, the columns of representation matrix $B$ are orthogonal to each other. $k$ is a positive constant which is the square of each column's norm.

The particularity for linear case is that, instead of optimizing two variables $B$ and $\bsyb{w}$, the algorithm actually only needs to learn the joint product $\bsyb{\theta}=B\bsyb{w}$, and the whole problem becomes $T$ linear regression on variables $\bsyb{\theta}^{(t)},t\in[T]$. At each level $h$, the algorithm gets $T$ solution vectors $\bsyb{\theta}^{(t)}_h,t\in[T]$. Take SVD for $\bsyb{\Theta}_h=[\bsyb{\theta}^{(1)}_h,\bsyb{\theta}^{(2)}_h,\hdots,\bsyb{\theta}^{(T)}_h]=U\Sigma V^{\top}$ and construct $\hat{B}_h$ as the $d$ top right vectors of $V$. Then optimize the best $\hat{\bsyb{w}}_{t,h}$ to let $\hat{B}_h \hat{\bsyb{w}}_{t,h}$ recover $\bsyb{\theta}^{(t)}_h$.

Based on the formulation above, our main result for linear representation is as below. 

\label{sec:thm1}
\textbf{Theorem 1} (main theorem for linear representations) \textit{
With any failure probability $\delta\in(0,1)$, under assumption 1, 2 and $2d\leq\min\{D,T\}$, if the sample size satisfies $N\gg \max\left\{D+\log\frac{HT}{\delta}, D^2 d^2\log( N)\right\}$ and $n\gg \kappa(d+\log\frac{H}{\delta})$, where $\kappa := \lambda_{min}^{-1}(\mathbb{E}_{x\sim \Mcal{D}^{(T+1)}}[\hat{B}_h^{\top} x x^{\top} \hat{B}_h])$, then with probability $1-\delta$ over the samples, the expected sub-optimality of the learned policy $\pi_{h}(s)=\arg\max_{a}\xi(s,a)^{\top} \hat{B}_h\hat{\bsyb{w}}_{T+1,h}$ for each state on the new task is bounded by}
\begin{align*}
   \|V^{\pi^{\star}}(s)-&V^{\pi}(s)\|\lesssim& (\sigma+1) H^2 \cdot  \sqrt{\frac{\kappa(d+\log(H/\delta))}{n}+\frac{D^2d^2\log( N)}{NT}+\frac{D(d^2+\log(H/\delta))}{N}}.
\end{align*}
We leave the detailed proof in appendix. An ideal $\Mcal{D}^{(T+1)}$ for target task should satisfy $\kappa = O(d)$. Then according to theorem 1 we know that it suffices to use only $n=O(\kappa d)=O(d^2)$ samples at each level to learn a good policy for target task with the learned representation. The complexity for learning from scratch without representation learning is simply setting $T=1$ and removing the term related to $n$, where the sub-optimality bound becomes $(\sigma+1)^2H^2\sqrt{\frac{D^2d^2 \log N}{N}}$ and $N=O(D^2d^2)$ is the required number of sample to learn a good policy for a single MDP without multitask representation learning. We can see it is much larger than $d^2$ by a factor of $D^2$.

\textbf{The importance of $\kappa$}. According to the definition of $\kappa$, the choice of $\mathcal{D}^{(T+1)}$ directly decides the magnitude of $\kappa$, which then affects the required sample size for $n$. Intuitively, $\kappa$ measures the expected number of sample to get enough information for the rarest appearing feature within distribution $\Mcal{D}^{(T+1)}$. In a challenging setting, some features are scarcely activated in the data sampled from $\Mcal{D}^{(T+1)}$, which means $\kappa$ can be up to $\Omega(D)$. Then the upper bound for required $n$ increases to $\Omega(Dd)$, which becomes much less useful.

\section{General Representation Function Analysis}
\label{sec:general}
In this section, we give bound for general non-linear function class of representation function $\phi(\cdot, \cdot)$. Analogous to the proof in linear cases, we need some other definitions and further assumptions to establish the bound for general representations.
These notions are all used in previous theoretical results in representation learning~\cite{du2020few}.
\begin{itemize}
    \item Define the \textbf{Gaussian width} for a set $S$ as 
    $$\Mcal{G}(S)=\mathbb{E}_{z\sim\Mcal{N}(0,I)}\sup_{v\in S}\langle v,z \rangle$$
    By Gaussian width, we can further define the complexity measure of a function class with respect to some specific input data $\Mcal{X}=\{X_t\}_{t=1}^{T}$
    \begin{align*}
        \Mcal{F}_{\Mcal{X}}(\Phi)=\{&A=[a_1,\hdots,a_T]\in\mathbb{R}^{N\times T}:\|A\|_F=1, \exists \phi,\phi'\in\Phi\quad s.t.\\ &a_t\in\operatorname{span}([\phi(X_t),\phi'(X_t)]),\forall t\in[T]\}
    \end{align*}
    
    \item Define \textbf{generalized covariance matrix} for two arbitrary representation functions $\phi,\phi'$ with respect to certain input distribution $q$ as
        $ \Sigma_q(\phi,\phi')=\mathbb{E}_{x\sim q}\left[\phi(x)\phi'(x)^{\top}\right]\in \mathbb{R}^{d\times d}$
    Also define the symmetric covariance matrix as 
    \begin{align*}
        \Lambda_q(\phi,\phi')
        &=\left[\begin{matrix}
        \Sigma_q(\phi,\phi)& \Sigma_q(\phi,\phi')\\
        \Sigma_q(\phi',\phi)& \Sigma_q(\phi',\phi')
        \end{matrix}\right]\in \mathbb{R}^{2d\times 2d} \succeq 0
    \end{align*}
    The divergence between two representations is 
    \begin{align*}
        &D_{q}\left(\phi, \phi^{\prime}\right):=\Sigma_{q}\left(\phi^{\prime}, \phi^{\prime}\right)-\Sigma_{q}\left(\phi^{\prime}, \phi\right)\left(\Sigma_{q}(\phi, \phi)\right)^{\dagger} \Sigma_{q}\left(\phi, \phi^{\prime}\right).
    \end{align*}
\end{itemize}
For general representation function class, the concentration for empirical covariance do not hold unconditionally. Hence we add two assumptions on top of the setting in linear representation.

\textbf{Assumption 6.1} (Point-wise concentration of covariance) \textit{For $\delta\in(0,1)$, there exists a number $N_{\operatorname{point}}(\Phi,p,\delta)$ such that if $N\geq N_{\operatorname{point}}(\Phi,p,\delta)$, then for any given $\phi,\phi'\in \Phi$, N i.i.d. samples of $p$ will with probability at least $1-\delta$ satisfy}
\begin{align*}
    0.9\Lambda_p(\phi,\phi')\preceq \Lambda_{\tilde{p}}(\phi,\phi') \preceq 1.1\Lambda_p(\phi,\phi')
\end{align*}
where $\tilde{p}$ is the empirical distribution uniform over the $N$ samples.

\textbf{Assumption 6.2} (Uniform concentration of covariance) \textit{For $\delta\in(0,1)$, there exists a number $N_{\operatorname{unif}}(\Phi,p,\delta)$ such that if $N\geq N_{\operatorname{unif}}(\Phi,p,\delta)$, then $N$ i.i.d. samples of $p$ will with probability at least $1-\delta$ satisfy}
\begin{align*}
    0.9\Lambda_p(\phi,\phi')\preceq \Lambda_{\tilde{p}}(\phi,\phi') \preceq 1.1\Lambda_p(\phi,\phi'),\quad \forall \phi,\phi'\in\Phi
\end{align*}

Typically, $N_{\operatorname{unif}}\gg N_{\operatorname{point}}$ because uniform concentration is a much stronger constraint. Usually, we expect $N_{\operatorname{unif}}=\tilde{O}(D)$ and $N_{\operatorname{point}}=\tilde{O}(d)$ as in section 5. Also, we suppose that the agent is aware of (or can compute) an ideal sampling distribution $q$ which satisfies the conditions below.

\textbf{Assumption 6.3} \textit{If two representation functions $\phi,\phi^{\prime}$ satisfy $\operatorname{Tr}[D_{q}\left(\phi, \phi^{\prime}\right)] \leq \epsilon $, then we know for any $\|x\|\leq 1$, $\|P_{\phi}^{\perp} \phi^{\prime}(x)\| \leq \epsilon$ holds, here $P_{\phi}^{\perp}=I-\Sigma_q(\phi,\phi)\Sigma_q(\phi,\phi)^{\dagger}$ means projection matrix onto orthogonal complement space of $\operatorname{span}(\Sigma_q(\phi,\phi))$.}

\textbf{Assumption 6.4} \textit{With respect to distribution $q$ defined above, for any $\phi,\phi^* \in \Phi$ and a small $\epsilon$ less than some constant threshold, if there exist $\bsyb{w},\bsyb{w}^*\in\mathbb{R}^d$ that satisfy}
\begin{align*}
    \mathbb{E}_{x\sim q} \| \phi(x)^{\top} \bsyb{w} - \phi^*(x)^{\top} \bsyb{w}^* \|^2 \leq \epsilon.
\end{align*}
\textit{Then there exist a constant invertible matrix $P$, such that }
\begin{align*}
    \| P\phi(x) - \phi^*(x) \|^2 = o(\epsilon/\|\bsyb{w}\|^2) = o(\epsilon/\|\bsyb{w}^*\|^2)  , \forall x: \|x\|\leq 1.
\end{align*}
 
Assumption 6.3 ensures the existence of an ideal distribution $q$ on which the learned representation can extrapolate, namely if in expectation $\operatorname{span}(\Sigma_q(\phi,\phi))$ and $\operatorname{span}(\Sigma_q (\phi',\phi'))$ are close, the for arbitrary input $x$ we know $\phi'(x)$ is close to $\operatorname{span}(\Sigma_q(\phi,\phi))$. Assumption 6.4 hypothesize the uniqueness for each $\phi$ in the sense of linear-transform equivalence. Two representation functions $\phi$ and $\phi'$ can yield similar estimation result if and only if they differ by just an invertible linear transformation. Note for the linear representation class $\Phi$, these assumptions are naturally satisfied. 

The main theorem for general representation function goes as follows

\textbf{Theorem 2} (Main Theorem for General Representations) \textit{For any failure probability $\delta \in (0,1)$, we use distribution $q$ for training and target task sampling. Denote $\kappa:=\lambda_{\min}^{-1}(\Sigma_q(\hat{\phi},\hat{\phi}))$. Suppose $N\geq N_{\operatorname{unif}}(\Phi,q,\frac{\delta}{3T}),N \gg \kappa\Mcal{G}(\Mcal{F}_{\Mcal{X}}(\Phi))^2 d$ and $n\geq N_{\operatorname{point}}(\Phi,q,\frac{\delta}{3}),n\gg \kappa d$. Under all assumptions above, then with probability at least $1-\delta$ over the samples, the sub-optimality of policy by predictor $\pi(s)= \arg\max_{a\in\Mcal{A}} \hat{\bsyb{w}}_{T+1}^{\top}\hat{\phi}(s,a)$ at first level on the target task suffers regret no more than}
\begin{align*}
  & \|V^{\pi^{\star}}(s)-V^{\pi}(s)\|\lesssim (\sigma+1) H^2 \cdot \sqrt{\frac{\Mcal{G}(\Mcal{F}_{\Mcal{X}}(\Phi))^2 \kappa d}{NT} + \frac{\kappa(d+\log(H/\delta))}{n}}.
\end{align*}
The detailed proof can be found in Appendix. The main structure of the proof is just an analogy to linear representation classes.

\begin{figure}[ht]
\centering
\label{fig:space}
\subfigure[]{
\includegraphics[width=0.36 \linewidth]{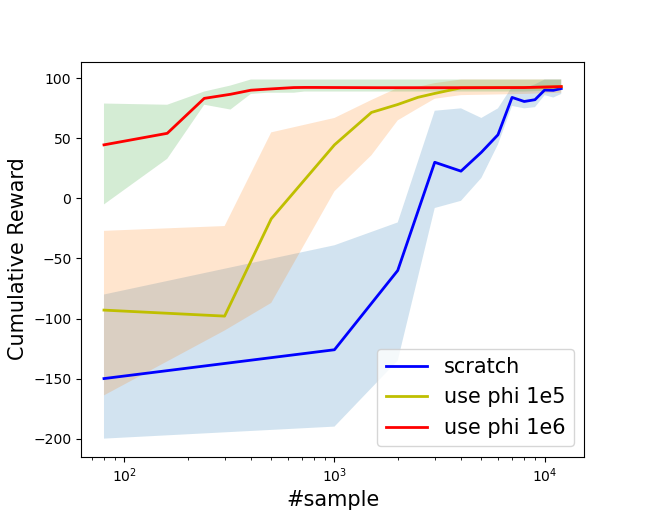}
}
\label{fig:expres}
\subfigure[]{
\includegraphics[width=0.6\linewidth]{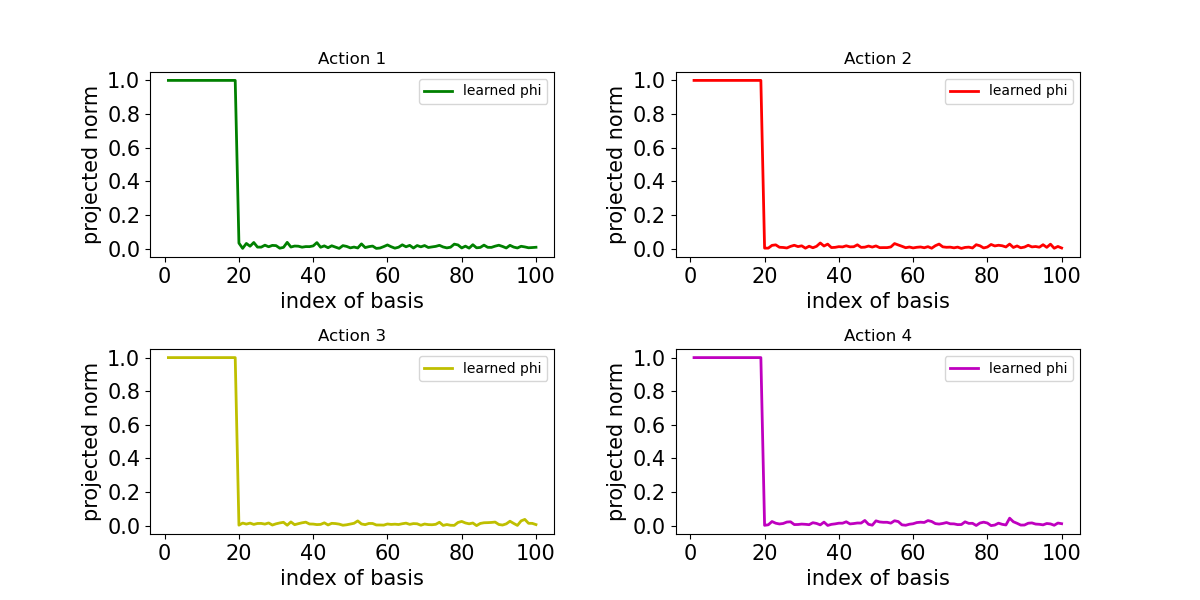}
}
\caption{One the left is the sample efficiency for naively learning from scratch (blue), learned representation $\hat{B}$ from $1e5$ samples per task (yellow), and $\hat{B}$ using $1e6$ samples (red). Notice that the x-axis is log-scale. On the right is the optimality of learned representation. The first 19 basis spans the optimal representation $B$ space. We can see from the projected norm $\|\hat{B}^{\top}_h e_i\|$ that the learned $\operatorname{span}(\hat{B}_h)$ is very close to ground truth $\operatorname{span}(B)$. We choose $\hat{B}_5$ learned from $1e6$ samples per task.
}
\end{figure}

\section{Empirical Experiment}
\label{sec:sec7}
\subsection{Environment}
We implement a specially designed noisy grid-world environment which permits a linear representation and \textit{satisfy all our assumptions} to conceptually verify our theoretical findings. The hidden state space is a set of discrete numbers $s_h\in\Mcal{S}=[K]$, of which $K$ is all possible positions of the agent. Each observed state $s_o$ consist of two parts $s_o=\left[o_1,o_2\right]^{\top}$, where $o_1=e_{s_h}\in\mathbb{R}^K$ is a one-hot vector denoting the position of the agent as $s_h$ and $o_2\sim \Mcal{N}(0,\sigma'^2 \mbf{I}_{D-K})$ is a purely random noise component. The action space contains four actions, namely up, down, left and right denoted as $\Mcal{A}=\{1,2,3,4\}$. There are three types of vacant positions: ground, fire and destination, each of them would return a reward of $-1,-10,+100$. After getting to destination the game is terminated.

Since the environment is essentially a tabular MDP, thus it must be linear MDP. An example of $\phi$ can be constructed as follows to verify that this grid-world environment permits a linear representation. First define $\xi(s_o,a)=[s_o^{\top}\delta_{a,1},s_o^{\top}\delta_{a,2},s_o^{\top}\delta_{a,3},s_o^{\top}\delta_{a,4}]$.
 Here $\delta_{i,j}=1$ if and only if $i=j$. So $\xi(\cdot,\cdot)$ essentially put observation into the block which corresponds to performed action. And the linear embedding function $B\in \mathbb{R}^{D\times d}$ can be constructed as 
$ B^{\star}=[e_1,\hdots,e_{K}, e_{D+1},\hdots,e_{D+K}, e_{2D+1},\hdots,e_{2D+K}, e_{3D+1},\hdots,e_{3D+K}] $
Intuitively, this $B$ filters out informationless entries and generate a compact representation $d=4K$. Composing those two steps we get a function $\phi(s_o,a)=(B^{\star})^{\top}\xi(s_o,a)$. 

\subsection{Implementation details}
We construct a grid-world size $5\times 5$ with 19 vacant positions, which means $K=19$. And set $D=100$ by adding $81$ redundant noisy dimensions of observation. We select 100 special state as barycentric coordinate basis which spans the whole observation space. Horizon is set to $H=10$ and it suffice to learn a good policy for most tasks. The multitask in this environment achieved by assigning different destination positions, fire configuration and action deviation probability $p$.

We sample $80$ different tasks, each one aims at individual destination position and has different action deviation probability $p$. Then we run algorithm 1 to generate learned weight $\hat{B}_h \hat{\bsyb{w}}_h\in \mathbb{R}^{400\times 80}$ for each level $h \in [H]$. Finally, we use this learned $\hat{B}$ to extract representation by $\hat{\phi}(x) = \hat{B}^{\top} x$ and run algorithm 2. To avoid singularity, we add $\lambda=0.01$ penalty for $\ell_2$ norm of for each $\hat{\bsyb{w}}_{T+1,h}$ as a regularization term (ridge regression).

We compare the performance of three different versions of learning procedure in a same evaluation environment with $p=0.05$, \textit{i) learning from scratch, namely set $T=1$ in algorithm 1, ii) using $\hat{B}$ which is learned from $N=10^5$ samples each task, iii) using $\hat{B}$ learned from $N=10^6$ samples}.
\subsection{Result}
To check to quality of our learned representation function $\hat{B}^{\top} x$, we quantify the distance between $\operatorname{span}(B)$ and $\operatorname{span}(\hat{B})$ by comparing the difference between $\|B^{\top} e_i\|$ and $\|\hat{B}_h^{\top} e_i\|$ for $i\in [D]$. This quantity measures $\|P_{\hat{B}_h^{\top}} B^{\top}\|$ by projecting each spanning basis of $B$ onto $\operatorname{span}(\hat{B}_h^{\top})$. We can see the learned representation function $\hat{B}_h$ is very close to $B$. (See \hyperref[fig:space]{Fig 1(b)}) 

Then we compare the performance of three different methods, including learning policy from scratch (set $T=1$ in algorithm 1), using learned representation $\hat{B}$ which is learned from different number of samples in training. As the figure illustrates, both experiments that use learned representation only needs small fraction of samples (around $3e2$ and $1e3$) compared with learning from scratch ($1e4$ sample). And the required sample size is even smaller when we use more samples to learn a better representation. 

\subsection{Verifying LAFA Criterion}
\begin{wrapfigure}{r}{0.5\textwidth}
\label{fig:fig2}
  \vspace{-30pt}
  \centering
    \includegraphics[width=0.48\textwidth]{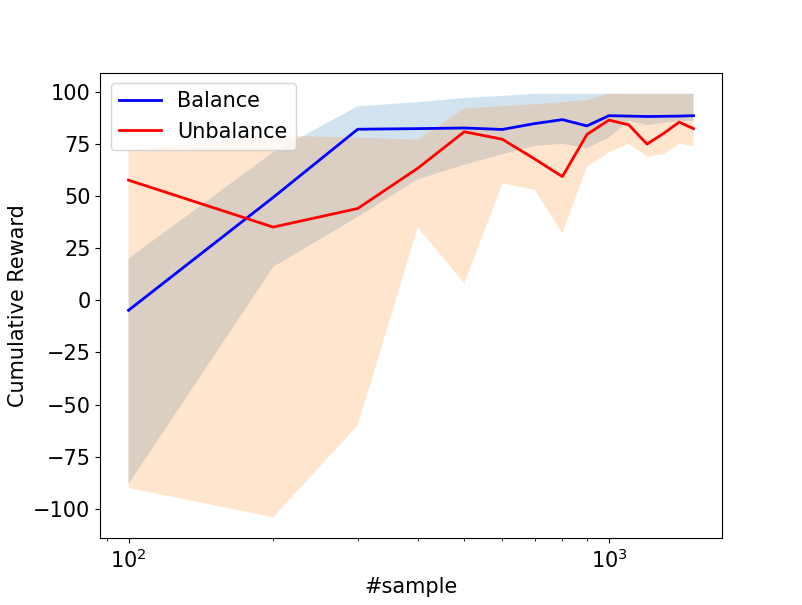}
  \caption{Performance curve for feature balanced distribution ($\kappa\approx 20$) and unbalanced distribution ($\kappa\approx 100$). The distribution whose feature abundance is unbalanced requires substantially more samples to stabilize.
  }
  \vspace{-10pt}
\end{wrapfigure}
In both theorems for linear and general representation cases, we introduced $\kappa$ to measure the abundance level for the least activated feature direction. We argued that the effect of reducing sample complexity for target new task is intimately related to $\kappa$. To check our finding, we construct two different sampling distributions. The algorithm samples natural basis $e_i\in\mathbb{R}^D$ and then refers to barycentric coordinate to get meta-data sample $x_i=\xi(s_i,a_i)$ among $\{(s_i,a_i)\}_{i=1}^D$. As for the balanced basis system, each hidden state (which is the position in maze) appears $D/d$ times in barycentric basis $s_i$, hence $\kappa=\lambda_{\min}^{-1}(\hat{B}^{\top}xx^{\top} \hat{B})=d$. While for unbalanced version, most of the hidden states $s_i$ correspond to the same position, i.e. starting position, and only one $s_j$ among all barycentric bases corresponds to each other position. This yields $\kappa=D$. 

Notice that we use ground truth representation function $\hat{B}=B$ for this experiment. We can see from figure \hyperref[fig:fig2]{2} that the unbalanced distribution with larger $\kappa$ requires around $1e3$ samples to achieve high performance stably, while for balanced distribution with smaller $\kappa$ it takes only $3e2$ samples to converge. This phenomenon demonstrates the fact that the sampling distribution greatly affects the sample complexity. When their LAFA criterion $\kappa$ is large, even the perfect representation function can be inefficient.

\section{Conclusion}
\label{sec:conclusion}
In this paper, we proposed a straightforward algorithm and gave a theoretical analysis demonstrating the power of multitask representation learning in linear MDPs. The algorithm utilize all the samples from different tasks to learn a representation which can generalize to novel unseen environment and accelerates the sample efficiency from $O(\mathcal{C}(\Phi)^2\kappa d H^4)$ to $O(\kappa d H^4)$. Here $\kappa$ is an important criterion called LAFA that depends on the distribution $\Mcal{D}$ used for learning new task. This demonstrates that a good representation alone is sometimes not sufficient to boost the sample efficiency if the sampling distribution used for learning policy is not well posed. 

\bibliographystyle{plain}
\bibliography{ref}
%%%%%%%%%%%%%%%%%%%%%%%%%%%%%%%%%%%%%%%%%%%%%%%%%%%%%%%%%%%%
\appendix
\newpage
\section*{Appendix}
\section{Proof Sketch}
To help reader better understand our proof, we first present a sketch of the overall procedure. The main structure of the proof consists of tow stages
\begin{itemize}
    \item Bounding the Q-value prediction error for each level $h$.
    \item Deriving the cumulative sub-optimality of the induced policy.
\end{itemize}

We overload the notation system used in our paper for this appendix. 
\subsection{Bounding Q-value Estimation Error in Each Level}
The goal for this stage is to give a uniform bound holds with probability $1-O(\delta)$ for $\forall x=\xi(s,a),s\in \Mcal{S},a\in \Mcal{A},t\in[T],h\in[H]$ as
\begin{align*}
    \left\| \hat{\bsyb{w}}_{t,h}^{\top}\hat{\phi}_h(x) - (\hat{\bsyb{w}}_{t,h}^{*})^{\top} \phi^*(x) \right\| \leq \text{something small}.
\end{align*}
Remind that $\xi(s,a)$ is a vectorization function encoding state-action pair as a $D$-dimensional vector. Here $\hat{\bsyb{w}}_{t,h}$ and $\hat{\phi}_{h}$ are solution to empirical optimization problem for task $t$ at level $h$. Similarly $\bsyb{w}^*_{t,h}$ and $\phi^*(x)$ correspond to ground truth weight vector and representation function. In order to bridge between learned Q-value function and authentic Q-value, we introduce $\hat{\bsyb{w}}_{t,h}^*$, which is defined by
$$ \hat{\bsyb{w}}_{t,h}^* = \theta_{t} + \sum_{s'\in\Mcal{S}} V_{h+1}^{(t)}(s')\cdot \psi^{(t)}(s') $$
Intuitively, this is the pseudo-ground truth weight vector induced by next level's value estimation $V_{h+1}^{(t)}(\cdot)$. Regarding $V_{h+1}^{(t)}(s)= \max_{a} \hat{\bsyb{w}}_{t,h+1}^{\top} \hat{\phi}_{h}(s,a)$ for every $s\in\Mcal{S}$ as perfect estimation, the best weight vector we can get at level $h$ for each task $t$ is such induced $ \hat{\bsyb{w}}_{t,h}^*$. 
We give a bound on the error of our learned parameter $\hat{\phi}_h(\cdot),\hat{\bsyb{w}}_{t,h}$ compared with $\phi_h^*(\cdot),\hat{\bsyb{w}}^*_{t,h}$ at each level $h$. Denote the sampled dataset for task $t$ as $X_t=[\xi(s_1^t,a_1^t),\hdots,\xi(s_n^t,a_n^t)]=[x_1^t,x_2^t,\hdots,x_n^t]\in \mathbb{R}^{D\times N}$ and $\phi(X_t)=[\phi(x_1^t),\phi(x_2^t),\hdots,\phi(x_n^t)]^{\top}$. The regression target value $\bsyb{y}_t$ for each $X_t$ is generated from
$$ \bsyb{y}_t = \phi^*(X_t) \hat{\bsyb{w}}^*_{t,h} + \bsyb{z}_{t,h}$$
Here $\bsyb{z}_{t,h}\in \mathbb{R}^N$ is the noise term, which is the sum of two noisy source $\bsyb{z}_{t,h}=\bsyb{z}_{t,h}^R+\bsyb{z}_{t,h}^T$. Every element of $\bsyb{z}_{t,h}^R$ is sampled from \textbf{R}eward noise distribution $\Mcal{Z}$, and $\bsyb{z}_{t,h}^T$ is the noise coming from finite sampling of \textbf{T}ransition dynamics

$$ \bsyb{z}^T_{t,h} \sim \frac{1}{m}\sum_{k=1}^m V^{(t)}_{h+1}(s'_{jk})-\mathbb{E}_{s'\sim P_{s_j^t,a_j^t}}V^{(t)}_{h+1}(s') $$
The first step is establishing in-sample error guarantee, we will prove that with probability at least $1-O(\delta)$
\begin{align*}
    \sum_{t=1}^T \left( \hat{\phi}_h(X_t) \hat{\bsyb{w}}_{t,h} - {\phi}_h^*(X_t) \hat{\bsyb{w}}^*_{t,h} \right)^2 \lesssim (\sigma+1)^2 \left( \Mcal{G}(\Mcal{F}_{\Mcal{X}}(\Phi))^2 + \log \frac{1}{\delta} \right)
\end{align*}
According to the condition that the number of sample is sufficiently large, we can get the expected excess risk's upper bound via covariance concentration
\begin{align*}
    \mathbb{E}_{x\sim \Mcal{D}^{(t)}} \left[ \left(\hat{\phi}_h(x) \hat{\bsyb{w}}_{t,h} - {\phi}^*(x) \hat{\bsyb{w}}^*_{t,h}\right)^2 \right] \lesssim \frac{1}{N} \left( \hat{\phi}_h(X_t) \hat{\bsyb{w}}_{t,h} - {\phi}^*(X_t) \hat{\bsyb{w}}^*_{t,h} \right)^2
\end{align*}
Together with conclusion that $\operatorname{span}(\hat{\phi}_h)$ is close to $\operatorname{span}(\phi^*)$, which is measured by projecting ground truth representation function $\phi^*(x)$'s output onto orthognal complement space of $\hat{\phi}(x)$
$$ \left\| P_{\hat{\phi}_h}^{\perp} \phi^*(x) \right\|^2 \leq \epsilon.$$
According to these results, we can finally establish the uniform Q-value prediction guarantee, such that 
$$ \left|\langle \hat{\phi}(s,a), \hat{\bsyb{w}}_h \rangle - \langle \phi^*(s,a), \hat{\bsyb{w}}^*_h \rangle \right| \leq \epsilon_h, \forall s\in\Mcal{S},a\in\Mcal{A},h\in[H] $$
\subsection{Deriving the Cumulative Sub-optimality of the Induced Policy}
Based on the error bound $\epsilon_h$ for arbitrary Q-value prediction at each level $h$ compared to pseudo ground truth as above, we deduce the upper bound for the sub-optimality of learned policy. We define several different value functions for each state $s$ at level $h$, including value induced by optimal policy $\pi^{\star}_h(s) = \arg\max_a \langle \phi^*(s,a), \bsyb{w}^*_{h} \rangle$ as $V^{\star}_h(s)$, value induced by learned policy $\pi_h(s) = \arg\max_a \langle \hat{\phi}(s,a), \hat{\bsyb{w}}_{h} \rangle$ as $V^{\pi}_h(s)$, value predicted by learned $\hat{\phi}(\cdot, \cdot), \hat{\bsyb{w}}_h$ as $\hat{V}^{\pi}_h(s)$ and value predicted by pseudo-optimal $\hat{\bsyb{w}}^*_h$ and $\phi^*(\cdot,\cdot)$ as $\tilde{V}^{\pi}_h(s) $.
\begin{align*}
    V^{\star}_h(s) :=& \mathbb{E}\left[ \sum_{k=h}^H r_{k}(s_h,\pi^{\star}(s_h))\Big| s_h=s \right],\\
    =& \max_a \langle \phi^*(s,a), \bsyb{w}^*_{h} \rangle ,\\
    V^{\pi}_h(s) :=& \mathbb{E}\left[ \sum_{k=h}^H r_{k}(s_h,\pi(s_h))\Big| s_h=s \right] ,\\
    \hat{V}^{\pi}_h(s) :=& \max_{a} \langle \hat{\phi}(s,a), \hat{\bsyb{w}}_h \rangle ,\\
    \tilde{V}^{\pi}_h(s) :=& \max_{a} \langle \phi^*(s,a), \hat{\bsyb{w}}^*_h \rangle.
\end{align*}

By cumulative error lemma (lemma A.4), we know the estimation value is close to optimal value, and real value is close to 
$$ \left| V^{\star}_h(s) - \hat{V}_h(s) \right| \leq  \sum_{k=h}^H \epsilon_k $$
Then we prove by induction that 
\begin{align*}
    &\left| V^{\pi}_h (s) - \hat{V}^{\pi}_h (s) \right| =\left| V^{\pi}_h (s) - \tilde{V}^{\pi}_h(s) + \tilde{V}^{\pi}_h(s) - \hat{V}^{\pi}_h (s) \right| \\
 \leq& \left| V^{\pi}_h (s) - \tilde{V}^{\pi}_h(s) \right| + \left| \tilde{V}^{\pi}_h(s) - \hat{V}^{\pi}_h (s) \right| \leq \epsilon_h + \max_{s} \left| V^{\star}_{h+1}(s) - \hat{V}_{h+1}(s) \right|\\
 \leq& \sum_{k=h}^H \epsilon_k.
\end{align*}
Hence we know
$$ \left| V^{\star}_h(s) - {V}^{\pi}_h(s) \right| \leq | V^{\star}_h(s) - {V}^{\pi}_h(s) | + | V^{\pi}_h(s) - \hat{V}_h(s) | \leq 2\sum_{k=h}^H \epsilon_k. $$
And this completes our proof.
\section{Appendix A. Linear Representation Class Analysis}
\label{linear_proof}
\subsection{Basic Lemmas}
To prove our main result \hyperref[sec:thm1]{Theorem 1}, we first prove several important lemmas. 

\textbf{Lemma A.1} \textit{Under the setting of theorem 1, with probability at least $1-\frac{\delta}{5}$ we have}
\begin{align*}
    \|\Mcal{X}(\hat{B}\hat{W}-B^*_h\hat{W}^*_h)\|_F^2\lesssim (\sigma+1)^2(Td+Dd\log(N)+\log(1/\delta)).
\end{align*}
\textit{Proof}. Let $\hat{\Theta}=\hat{B}\hat{W}$ and $\Theta^*=B^*_h\hat{W}^*_h$, $\Delta:=\hat{\Theta}-\Theta^*$. From the empirical optimality we have $\|Y-\Mcal{X}(\hat{\Theta})\|_F^2\leq \|Y-\Mcal{X}(\Theta^*)\|_F^2$. Plugging in $Y=\Mcal{X}(\Theta^*)+Z$ and get 
\begin{align}
    \|\Mcal{X}(\hat{\Theta}-\Theta^*)\|_F^2\leq 2 \langle Z,\Mcal{X}(\hat{\Theta}-\Theta^*)\rangle.
\end{align}
Since $\Theta^*$ and $\hat{\Theta}$ can be both factored into two matrices whose rank is no larger than $d$, we have $\operatorname{rank}(\hat{\Theta}-\Theta^*)\leq 2d$. Thus
we can factor $\Delta=VR=[Vr_1,Vr_2,\hdots,Vr_T]$ where $V\in \Mcal{O}_{D,2d}=\{A\in\mathbb{R}^{D\times 2d}:A^{\top}A=I\}$ and $R=[r_1,r_2,\hdots,r_T]$. By this transformation, we factor $\operatorname{span}(\Delta)$ into orthonormal basis. For each $t\in[T]$, denote $X_tV=U_tQ_t$ where $U_t\in \Mcal{O}_{N,2d}$ and $Q_t\in \mathbb{R}^{2d\times 2d}$. Then we have
\begin{align*}
    \langle Z,\Mcal{X}(\Delta)\rangle
    &=\sum_{t=1}^T z_t^{\top} X_tVr_t\\
    &=\sum_{t=1}^T z_t^{\top}U_tQ_tr_t\\
    &\leq \sqrt{\sum_{t=1}^T \|U_t^{\top}z_t\|^2}\cdot \sqrt{\sum_{t=1}^T \|Q_tr_t\|^2}\\
    &=\sqrt{\sum_{t=1}^T \|U_t^{\top}z_t\|^2}\cdot \sqrt{\sum_{t=1}^T \|X_tVr_t\|^2}\\
    &=\sqrt{\sum_{t=1}^T\|U_t^{\top}z_t\|^2}\cdot\|\Mcal{X}(\Delta)\|_F\quad (*)
\end{align*}
Then we will give a high-probability upper bound on both $\sum_{t=1}^T \|U_t^{\top}z_t\|^2$ and $\|\Mcal{X}(\Delta)\|_F$. 

According to (1), we know $\|\Mcal{X}(\Delta)\|_F^2\leq 2\|Z\|_F\|\Mcal{X}(\Delta)\|_F$, which indicates that $\|\Mcal{X}(\Delta)\|_F \leq 2\|Z\|_F\lesssim (\sigma+1)\sqrt{N T+\log(1/\delta)}$ by technical lemma B.3 (a loose bound). Denote $\Delta=[\delta_1,\delta_2,\hdots,\delta_T]$, then we have the relationship for $\|\Delta\|_F^2$ and $\|\Mcal{X}(\Delta)\|$
\begin{align*}
    \|\Mcal{X}(\Delta)\|_F^2&=\sum_{t=1}^{T}\|X_t\delta_t\|^2\\
    &=\sum_{t=1}^T \delta_t^{\top}X_t^{\top}X_t \delta_t\\
    &\geq 0.9 N\sum_{t=1}^T \delta_t^{\top}\Sigma \delta_t\\
    &\geq 0.9 N \sum_{t=1}^T \lambda_{\min}(\Sigma)\|\delta_t\|^2\\
    &=0.9 N \underline{\lambda} \|\Delta\|_F^2.
\end{align*}
where $\underline{\lambda}=\lambda_{\min}(\Sigma)$. Hence we get the bound for $\|\Delta\|_F^2$ as 
\begin{align*}
    \|\Delta\|_{F}^{2} \lesssim \frac{\|\mathcal{X}(\Delta)\|_{F}^{2}}{N \lambda} \lesssim \frac{\sigma^{2}\left(N T+\log (1 / \delta)\right)}{N \underline{\lambda}}.
\end{align*}
Notice that $U_t$ depends on $V$, which is learned from $X_t$ and $Y_t$, therefore further depends on $Z$. So independent concentration bound cannot be applied directly. Instead, we will use $\epsilon$-net argument to establish the bound for $\|\Mcal{X}(\Delta)\|$. For any fixed $\bar{V}\in\Mcal{O}_{D,2d}$, let $X_T\bar{V}=\bar{U}_t\bar{Q}_t$ where $\bar{U}_t\in\Mcal{O}_{N,2d}$. Since each $\bar{U}_t^{\top}z_t$ has at most $2d$ degree of freedom, by technical lemma B.1, we know with probability at least $1-\delta'$ over $Z$
$$\sum_{t=1}^T\|\bar{U}_t^{\top}z_t\|^2\lesssim (\sigma+1)^2(Td+\log(1/\delta'))$$

Plug into the inequality as $(*)$ where $\Delta=\bar{V}R$, we have
\begin{align*}
    \langle Z,\Mcal{X}(\bar{V}R)\rangle\lesssim (\sigma+1)\sqrt{Td+\log(1/\delta')}\|\Mcal{X}(\bar{V}R)\|_F.
\end{align*}

By technical lemma B.6 we know there exist an $\epsilon$-net $\Mcal{N}$ of $\Mcal{O}_{D,2d}$ and the cardinality satisfies $|\Mcal{N}|\leq \left(\frac{6\sqrt{2d}}{\epsilon}\right)^{2Dd}$. Then by union bound, we know that with probability at least $1-\delta'|\Mcal{N}|$,
\label{equ:eq2}
\begin{align}
    \langle Z,\Mcal{X}({V}R)\rangle\lesssim (\sigma+1)\sqrt{Td+\log(1/\delta')}\|\Mcal{X}({V}R)\|_F,\quad \forall{V}\in\Mcal{N}.
\end{align}
Choose $\delta'=\frac{\delta}{20}\left(\frac{\epsilon}{6\sqrt{2}d}\right)^{2Dd}$, so we know (2) holds with probability at least $1-\delta/20$. Next we apply this $\epsilon$-net to bound $\|\Mcal{X}(\Delta)\|_F^2$, for any $V$, by technical lemma B.6 we know there exists a $\bar{V}\in \Mcal{N}$ such that $\|V-\bar{V}\|_F \leq \epsilon$
\begin{align*}
\|\Mcal{X}(\Delta)\|_F^2=&\|\Mcal{X}(VR-\bar{V}R)\|_F^2\\
     =&\sum_{t=1}^{T} \|X_t(V-\bar{V})r_t\|^2\\
  \leq&\sum_{t=1}^T\|X_t\|^2 \cdot \|V-\bar{V}\|^2 \cdot \|r_t\|^2\\
  \leq&\sum_{t=1}^T 1.1n\lambda_{\max}(\Sigma)\epsilon^2\|r_t\|^2\\
  \leq&1.1n\bar{\lambda}\epsilon^2\|R\|_F^2\\
     =&1.1n\bar{\lambda}\epsilon^2\|\Delta\|_F^2\\
 \lesssim&N\bar{\lambda}\epsilon^2\cdot\frac{(\sigma+1)^2(N T+\log(1/\delta))}{N\underline{\lambda}}\\
 =&\epsilon^2 \sigma^2(N T+\log(1/\delta))    \tag{(2a), $\bar{\lambda} / \underline{\lambda}=1$}
\end{align*}

Finally we can finish the proof by selecting a proper $\epsilon$ and corresponding $\delta'$ to obtain the best trade-off.
\begin{align*}
    &\frac{1}{2}\|\Mcal{X}(\Delta)\|_F^2\\
  \leq&\langle Z,\Mcal{X}(\Delta)\rangle\\
  =&\langle Z,\Mcal{X}(\bar{V}R)\rangle+\langle Z,\Mcal{X}(VR-\bar{V}R)\rangle\\
 \lesssim& (\sigma+1)\sqrt{Td+\log(1/\delta')}\|\Mcal{X}(\bar{V}R)\|_F+\|Z\|_F \|\Mcal{X}(VR-\bar{V}R)\|_F\\
  \leq& (\sigma+1)\sqrt{Td+\log(1/\delta')}(\|\Mcal{X}(VR)\|_F+\|\Mcal{X}(\bar{V}R-VR)\|_F)\\
  &+(\sigma+1)\sqrt{N T+\log(1/\delta')} \|\Mcal{X}(VR-\bar{V}R)\|_F\\
  \lesssim& (\sigma+1)\sqrt{Td+\log(1/\delta')}\|\Mcal{X}(VR)\|_F+(\sigma+1)\sqrt{N T+\log(1/\delta')} \|\Mcal{X}(VR-\bar{V}R)\|_F.
\end{align*}
The last inequality holds because $d<N$ and $\delta'<\delta$. Plug in the bound of (2a), we know
\begin{align*}
    &\frac{1}{2}\|\Mcal{X}(\Delta)\|_F^2\\
 \lesssim& (\sigma+1)\sqrt{Td+\log(1/\delta')}\|\Mcal{X}(\Delta)\|_F+(\sigma+1)\sqrt{N T+\log(1/\delta')}\sqrt{\epsilon^2 \sigma^2(N T+\log(1/\delta))}\\
 \leq& (\sigma+1)\sqrt{Td+\log(1/\delta')}\|\Mcal{X}(\Delta)\|_F+\epsilon(\sigma+1)^2( N T+\log(1/\delta')).
\end{align*}
Denote $B_1=(\sigma+1)\sqrt{Td+\log(1/\delta')}$ and $B_2=\sqrt{\epsilon (\sigma+1)^2(N T+\log(1/\delta'))}$, then the inequality above can be represented in an equivalent way as
\begin{align*}
    \frac{1}{2}\|\Mcal{X}(\Delta)\|_F^2-B_1\|\Mcal{X}(\Delta)\|&\leq B_2^2\\
    \frac{1}{2}(\|\Mcal{X}(\Delta)\|_F -B_1)^2 &\leq \frac{1}{2}B_1^2+B_2^2 \lesssim \max\{B_1^2,B_2^2\}\\
    (\|\Mcal{X}(\Delta)\|_F-B_1)^2 &\lesssim \max\{B_1,B_2\}\\
    \|\Mcal{X}(\Delta)\|_F &\lesssim B_1+\max\{B_1,B_2\}\lesssim \max\{B_1,B_2\}
\end{align*}
Finally we set $\epsilon=\frac{d}{N}$ and corresponding $\delta'=\frac{\delta}{20}\left(\frac{\epsilon}{6\sqrt{2d}}\right)^{2Dd}$, so that 
\begin{align*}
    B_2&=(\sigma+1)\sqrt{\frac{d}{N}(N T+\log(1/\delta'))}\\
    &\leq(\sigma+1)\sqrt{Td+\log(1/\delta')}\\
    &=B_1
\end{align*}
This gives the final bound for $\|\Mcal{X}(\Delta)\|_F$
\begin{align*}
    \|\Mcal{X}(\Delta)\|_F\lesssim& (\sigma+1)\sqrt{Td+\log(1/\delta')}\\
    \lesssim& (\sigma+1)\sqrt{Td+Dd\log\frac{d}{\epsilon}+\log\frac{1}{\delta}}\\
    \leq&(\sigma+1)\sqrt{Td+Dd\log(N)+\log\frac{1}{\delta}}
\end{align*}
Take square and complete the proof.
\qed
\\

\textbf{Lemma A.2} (Feature Space Distance Guarantee) \textit{Under assumptions described in section 6, with probability at least $1-\frac{2\delta}{5}$, we have}
\begin{align*}
     &\left\|P_{\Sigma^{1 / 2}\hat{B}_h}^{\perp} \Sigma^{1 / 2} \hat{B}^{*}_h \right\|_{F}^{2}
    \lesssim\frac{(\sigma+1)^2(Td+Dd\log( N)+\log\frac{1}{\delta})}{N\sigma_d^2(\hat{W}^*_h)}
\end{align*}
\textit{Proof}
By lemma A.1 we have
\begin{align*}
    &(\sigma+1)^2(Td+Dd\log(N)+\log(1/\delta))\\
    \gtrsim&\|\Mcal{X}(\hat{B}_h\hat{W}_h - \hat{B}^*_{h} \hat{W}^*_h)\|_F^2\\
    =&\sum_{t=1}^T \left\| X_t \hat{B} \hat{\bsyb{w}}_{t,h} -  X_t \hat{B}^*_h \hat{\bsyb{w}}_{t,h}^* \right\|^2\\
    =&\sum_{t=1}^{T}\left\|P_{X_{t} \hat{B}}\left(X_{t} \hat{B}^{*}_h \hat{\bsyb{w}}_{t,h}^{*}+z_{t}\right)-X_{t} \hat{B}^*_{h} \hat{w}_{t}^{*}\right\|^{2} \\
    =&\sum_{t=1}^{T}\left\|-P_{X_{t} \hat{B}_h }^{\perp} X_{t} \hat{B}^*_{h} \hat{\bsyb{w}}_{t,h}^{*}+P_{X_{t} \hat{B}_h} z_{t}\right\|^{2} \\
    =&\sum_{t=1}^{T}\left(\left\|-P_{X_{t}\hat{B}}^{\perp} X_{t}^{*} \hat{B}^*_{h} \hat{\bsyb{w}}_{t,h}^{*}\right\|^{2}+\left\|P_{X_{t} \hat{B}_h} z_{t}\right\|^{2}\right) \\
    \geq&\sum_{t=1}^{T}\left\|P_{X_{t}\hat{B}}^{\perp} X_{t} \hat{B}^*_{h} \hat{\bsyb{w}}_{t,h}^{*}\right\|^{2} \\
    \geq&0.9 N \sum_{t=1}^{T}\left\|P_{\Sigma^{1 / 2}\hat{B}}^{\perp} \Sigma^{1 / 2} \hat{B}^*_{h} \hat{\bsyb{w}}_{t,h}^{*}\right\|^{2} \tag{Technical lemma B.2 and B.4} \\
    =&0.9 N\left\|P_{\Sigma^{1 / 2}\hat{B}}^{\perp} \Sigma^{1 / 2} \hat{B}^*_{h} \hat{W}^{*}_h \right\|_{F}^{2} \\
    \geq&0.9 N\left\|P_{\Sigma^{1 / 2}\hat{B}}^{\perp} \Sigma^{1 / 2} \hat{B}^*_{h}\right\|_{F}^{2} \cdot \sigma^{2}_d \left(\hat{W}^{*}_h \right).
\end{align*}
\qed
\\

\textbf{Lemma A.3} (Similarity guarantee for learned vector) \textit{If $N \gg D^2 d^2 \log(N)$ for every level $h \in [H]$, then we have }
\begin{align*}
    \left| \| \hat{\bsyb{w}}^*_{t,h}\| - \|\bsyb{w}_{t,h}^* \| \right| &= o\left(\frac{H-h}{d}\right),\quad for\ \forall h\in[H], t\in[T].
\end{align*}
For target task, in addition to constraint above, we also need $n \gg d^2$ to ensure that
\begin{align*}
    \left| \|\bsyb{w}_{T+1,h}^*\| - \|\hat{\bsyb{w}}_{T+1,h}\| \right| =  o(H-h+1).
\end{align*}
\textit{proof}.
We just need to show that under given condition we have
$$\| \hat{B}_h^*\hat{\bsyb{w}}^*_{t,h} - B^* \bsyb{w}^*_{t,h}\| = o\left(\frac{H-h}{d}\right), $$
then we will know $ \left| \| \hat{\bsyb{w}}^*_{t,h}\| - \|\bsyb{w}_{t,h}^* \| \right| = \left| \| \hat{B}_h^* \hat{\bsyb{w}}^*_{t,h}\| - \|B^* \bsyb{w}^*_{t,h}\|\right| \leq \| \hat{B}_h^* \hat{\bsyb{w}}^*_{t,h} - B^* \bsyb{w}^*_{t,h}\| = o\left(\frac{H-h}{d}\right)$. Setting $T=1$ in lemma A.1, we know for any level $h \in [H]$,
\begin{align*}
    &\mathbb{E}_{x\sim p_x}[x^{\top}(\hat{B}_{h} \hat{\bsyb{w}}_{t,h} - \hat{B}^*_{h} \hat{\bsyb{w}}^*_{t,h})]^2 \\
    = & (\hat{B}_{h}\bsyb{w}_{t,h} - \hat{B}^*_{h} \bsyb{w}^*_{t,h})^{\top} \Sigma (\hat{B}_{h} \hat{\bsyb{w}}_{t,h} - \hat{B}^*_{h} \hat{\bsyb{w}}^*_{t,h}) \\
    \leq& \frac{1}{0.9 N} (\hat{B}_{h} \hat{\bsyb{w}}_{t,h} - \hat{B}^*_{h} \hat{\bsyb{w}}^*_{t,h} )^{\top} X_t X_t^{\top} (\hat{B}_{h} \hat{\bsyb{w}}_{t,h} - \hat{B}^*_{h} \hat{\bsyb{w}}^*_{t,h}) \\
    =& \frac{1}{0.9 N}\left\| X_t (\hat{B}_{h} \hat{\bsyb{w}}_{t,h} - \hat{B}^*_{h} \hat{\bsyb{w}}^*_{t,h})\right\|^2 \\
    \lesssim& \frac{(\sigma+1)^2\left( Dd\log(N) + d + \log(1/\delta) \right)}{N}
\end{align*}
From sample distribution we know $\Sigma=\mathbb{E}_{x\sim p_x}[x x^{\top}]=I_D/D$, hence we know $(\hat{B}_{h}\bsyb{w}_{t,h} - \hat{B}^*_{h} \bsyb{w}^*_{t,h})^{\top} \Sigma (\hat{B}_{h} \hat{\bsyb{w}}_{t,h} - \hat{B}^*_{h} \hat{\bsyb{w}}^*_{t,h}) = \frac{1}{D} \|\hat{B}_{h}\bsyb{w}_{t,h} - \hat{B}^*_{h} \bsyb{w}^*_{t,h} \|^2$. We can further get uniform guarantee of training data. Namely, for any $\|x\| \leq 1$, we have
\begin{align*}
    &\left\| x^{\top} (\hat{B}_{h} \hat{\bsyb{w}}_{t,h} - \hat{B}^*_{h} \hat{\bsyb{w}}^*_{t,h}) \right\|^2 \\
    \leq &  \left\| \hat{B}_{h} \hat{\bsyb{w}}_{t,h} - \hat{B}^*_{h} \hat{\bsyb{w}}^*_{t,h} \right\|^2 \\
    \leq  &  D\cdot \frac{(\sigma+1)^2\left( Dd\log(N) + d + \log(1/\delta) \right)}{N} \\
    = & o\left( \frac{1}{d}\right)
\end{align*}
By lemma A.4, plug in $V^{\star}(s)=\max_{a\in\Mcal{A}} \xi(s,a)^{\top} B^*\bsyb{w}^*_{t,h}$ as anchor value and $\hat{V}_h(s)=\max_{a\in\Mcal{A}} \xi(s,a)^{\top} \hat{B}_h \bsyb{w}_{t,h}$, we know $\left| V^{\star}_h(s) - \hat{V}_h(s) \right| = o\left( \frac{H-h+1}{d} \right)$.

Finally, consider the difference of $B^* \bsyb{w}^*_{t,h}$ and $\hat{B}^*_h \hat{\bsyb{w}}^*_{t,h}$ for any $x=\xi(s,a)$. From the definition of linear MDP, we can set $\hat{B}_h^*=B^*$ (other equivalent $\hat{B}^*$ which satisfies $\operatorname{span}(\hat{B}^*_h)=\operatorname{span}(B^*)$ does not affect $\hat{\bsyb{w}}^*_{h,t}$'s norm, and can be reduced to our setting by $\hat{\bsyb{w}}^*_{t,h} \mapsto (B^*)^{\top} \hat{B}^*_h \hat{\bsyb{w}}^*_{t,h}, \hat{B}^*_h \mapsto B^*$). Then we know $\hat{\bsyb{w}}^*_{t,h}$ and $\bsyb{w}^*_{t,h}$ is generated from equation
\begin{align*}
    \hat{\bsyb{w}}^*_{t,h} &= \theta + \sum_{s'} \hat{V}_{h+1}(s')\cdot \psi(s')\\
    \bsyb{w}^*_{t,h} &= \theta + \sum_{s'} \hat{V}_{h+1}(s') \cdot \psi(s')
\end{align*}
Hence we know 
\begin{align*}
   &\left\| B^*\hat{\bsyb{w}}^*_{t,h} - B^*\bsyb{w}^*_{t,h} \right\| \\
 = &\left\|B^* \sum_{s'} \psi(s') [V^{\star}_{h+1}(s')-\hat{V}_{h+1}(s') \right\| \\
 \leq& \max_{s\in\Mcal{S}} \left| V^{\star}_{h+1}(s) - \hat{V}_{h+1}(s) \right|\cdot \left| \sum_{s'\in \Mcal{S}} B^*\psi(s') \right|  \\
 =& \max_{s\in\Mcal{S}} \left| V^{\star}_{h+1}(s) - \hat{V}_{h+1}(s) \right| \\
 =&o\left(\frac{H-h}{d}\right).
\end{align*}
And this gives the first result in our lemma. Notice the whole logic can be summarized as $\| \hat{\bsyb{w}}^*_{t,h} - \bsyb{w}^*_{t,h} \| \leq \sum_{k=h+1}^{H} \delta_k$ where $\delta_k$ is the prediction error of $\hat{B}_{k} \hat{\bsyb{w}}_{t,k}$ compared with $\hat{B}^*_{k} \hat{\bsyb{w}}^*_{t,k}$. Hence, when $n \gg d^2$, we know from main theorem that $\delta_k = o(1)$, so $\left| \|\bsyb{w}_{T+1,h}^*\| - \|\hat{\bsyb{w}}_{T+1,h}\| \right| =  o(H-h+1)$.
\qed
\\

\textbf{Lemma A.4} (Value prediction error accumulation) \textit{Suppose we have one perfect anchor value function $V^*_h(s)$ and another deviated Q-value estimation $Q_h(s,a)$ for $h\in[H]$. Notice $Q_h(s,a)$ based its estimation on next level $\hat{V}_{h+1}(s)=\max_a Q_{h+1}(s,a)$. If at level $h\in [H]$, the Q-value estimation error for $Q_h(s,a)$ is bounded by $\delta_h$, namely}
\begin{align*}
    \left| Q_h(s,a) - r_h(s,a) - [\Mcal{T}_a \hat{V}_{h+1}](s)\right| &\leq \delta_h,\quad for\ \forall (s,a)\in\Mcal{S}\times\Mcal{A}.\\
    \hat{V}_{H+1}(s)&=V^*_{H+1}(s)=0
\end{align*}
\textit{Then we have for any $s\in \Mcal{S}$ and level $h\in[H]$}
\begin{align*}
    \left| V^*_h(s) - \hat{V}_h(s) \right| \leq \sum_{k=h}^H \epsilon_h
\end{align*}
\textit{Proof.} We prove it recursively by math induction. For level $H+1$ we know the claim holds. Now suppose it holds for level $h+1$. Denote $\hat{a}=\arg\max_{a} Q_h(s,a)$ and $a=\arg\max_{a}Q^*_h(s,a)= \arg\max_{a} r_h(s,a)+[\Mcal{T}_a V^*_{h+1}](s)$. If $V^*_h(s) - Q_h(s,a) > 0$, then we have
\begin{align*}
     &V^*_h(s) - \hat{V}_h(s)\\
    =&V^*_h(s) - Q_h(s,\hat{a})\\
    \leq & V^*_h(s) - Q_h(s,\hat{a}) + (Q_h(s,\hat{a}) - Q_h(s,a)) \tag{optimality of $\hat{a}$ for $Q_h(s,a)$}\\
    =& r_h(s,a)+[\Mcal{T}_a V^*_{h+1}](s) - Q_h(s,a) \\
    =& r_h(s,a)+[\Mcal{T}_a V^*_{h+1}](s) - r_h(s,a) - [\Mcal{T}_a \hat{V}_{h+1}](s) - (Q_h(s,a) - r_h(s,a) - [\Mcal{T}_a \hat{V}_{h+1}](s)) \\
    =& [\Mcal{T}_a (V^*_{h+1}-\hat{V}_{h+1})](s) - (Q_h(s,a) - r_h(s,a) - [\Mcal{T}_a \hat{V}_{h+1}](s)).
\end{align*}

otherwise we know $V^*_h(s) - Q_h(s,a) < 0$, then we have
\begin{align*}
     &\hat{V}_h(s) - V^*_h(s)\\
    =&Q_h(s,\hat{a}) - V^*_h(s)\\
    \leq& Q_h(s,\hat{a}) - V^*_h(s) + (Q^*_h(s,a) - Q^*_h(s,\hat{a})) \tag{optimality of $a$ for $Q^*_h(s,a)$}\\
    =& Q_h(s,\hat{a}) - (r_h(s,a)+[\Mcal{T}_a V^*_{h+1}](s))\\
    =& r_h(s,\hat{a}) + [\Mcal{T}_{\hat{a}} \hat{V}_{h+1}](s) - r_h(s,\hat{a})+[\Mcal{T}_{\hat{a}} V^*_{h+1}](s) + (Q_h(s,\hat{a}) - r_h(s,\hat{a}) - [\Mcal{T}_{\hat{a}} \hat{V}_{h+1}](s)) \\
    =& [\Mcal{T}_a (\hat{V}_{h+1}-V^*_{h+1})](s) + (Q_h(s,\hat{a}) - r_h(s,\hat{a}) - [\Mcal{T}_{\hat{a}} \hat{V}_{h+1}](s)).
\end{align*}
We know by given condition that, for any $(s,a) \in \Mcal{S}\times\Mcal{A}$
\begin{align*}
    \left| Q_h(s,a) - r_h(s,a) - [\Mcal{T}_a \hat{V}_{h+1}](s)) \right| \leq \epsilon_h.
\end{align*}
By induction result, we know for any $(s,a) \in \Mcal{S}\times\Mcal{A}$
\begin{align*}
    \left| [\Mcal{T}_a (V^*_{h+1}-\hat{V}_{h+1})](s) \right| &\leq \max_{s\in\Mcal{S}} \left| V^*_{h+1}(s)-\hat{V}_{h+1}(s) \right| \leq \sum_{k=h+1}^H \delta_k.
\end{align*}
Adding together, we indeed verify that  $\left| V^*_h(s) - \hat{V}_h(s) \right| \leq \sum_{k=h}^H \delta_h$ holds for $h$. Hence by induction we know the lemma holds for any $h\in[H]$.
\qed
\\

%%\textbf{Lemma A.} 
Next we will prove key lemma.

\textbf{Key Lemma.} \textit{Denote $\Sigma_t=\mathbb{E}_{x\sim \mu} \left[\hat{B}^{\top} x x^{\top} \hat{B}\right]$. If we have expected error bound within distribution $\mu$ in adaptive sampling is bounded by}
\begin{align*}
    \frac{1}{2} \mathbb{E}_{x\sim \mu}\left[ x^{\top} (\hat{B} 
\hat{\bsyb{w}} - \hat{B}^* \hat{\bsyb{w}}^* ) \right]^2\leq \zeta_1,
\end{align*}
and meanwhile 
\begin{align*}
    \left\|P_{\hat{B}}^{\perp} \hat{B}^{*} \right\|_{F}^{2}\leq \zeta_2,
\end{align*}
Then we know for any $x,\|x\|\leq 1$
\begin{align*}
    \left\| x^{\top}(\hat{B}\hat{\bsyb{w}} - \hat{B}^* \hat{\bsyb{w}}^* ) \right\| \lesssim \sqrt{\lambda_{\min}^{-1}(\Sigma_t)\zeta_1} + \| \hat{w}^* \| \sqrt{\zeta_2 }.
\end{align*}
\textit{proof.} For any $\|x\|\leq 1$, decompose it into $x=P_{\hat{B}}x+P_{\hat{B}}^{\perp} x$. Then we have
\begin{align*}
     \left\| x^{\top}(\hat{B}\hat{\bsyb{w}} - \hat{B}^* \hat{\bsyb{w}}^* ) \right\|&\leq  \left\| x^{\top} P_{\hat{B}} (\hat{B}\hat{\bsyb{w}} - \hat{B}^* \hat{\bsyb{w}}^* ) \right\| + \left\| x^{\top} P_{\hat{B}}^{\perp} (\hat{B}\hat{\bsyb{w}} - \hat{B}^* \hat{\bsyb{w}}^* ) \right\| \tag{$\star$}
\end{align*}
The second term in $(\star)$ can be calculated from
\begin{align*}
    &\left\| x^{\top} P_{\hat{B}}^{\perp} (\hat{B}\hat{\bsyb{w}} - \hat{B}^* \hat{\bsyb{w}}^* ) \right\|^2 \\
    =&\left\| x^{\top} P_{\hat{B}}^{\perp} \hat{B}\hat{\bsyb{w}} - x^{\top} P_{\hat{B}}^{\perp} \hat{B}^* \hat{\bsyb{w}}^*  \right\|^2 \\
    =&\left\| x^{\top} P_{\hat{B}}^{\perp} \hat{B}^* \hat{\bsyb{w}}^*  \right\|^2 \\
    \leq & \|x\|^2 \cdot \|P_{\hat{B}}^{\perp} \hat{B}^*\|^2 \cdot \|\hat{\bsyb{w}}^*\|^2
\end{align*}
By our assumption we know $\|x\|^2\leq 1$ and $\|P_{\hat{B}}^{\perp} \hat{B}^*\|^2 \leq \|P_{\hat{B}}^{\perp} \hat{B}^*\|_F^2 = \zeta_2$, hence we have 
\begin{align*}
    \left\| x^{\top} P_{\hat{B}}^{\perp} (\hat{B}\hat{\bsyb{w}} - \hat{B}^* \hat{\bsyb{w}}^* ) \right\|^2 \leq \zeta_2 \| \hat{\bsyb{w}}^* \|^2 \tag{a}
\end{align*} 
Then we give bound for the first term of $(\star)$. Notice that $P_{\hat{B}} \hat{B}^* = \hat{B} \hat{B}^{\dagger} \hat{B}^*:= \hat{B} P$ where $P=\hat{B}^{\dagger} \hat{B}^* \in \mathbb{R}^{d \times d}$. Then we have
\begin{align*}
    & \mathbb{E}_{x\sim \mu} \left\| x^{\top} \hat{B} (\hat{\bsyb{w}}-P\hat{\bsyb{w}}^*) \right\|^2 \tag{b} \\
   =& \mathbb{E}_{x\sim \mu} \left\| x^{\top} (\hat{B} \hat{\bsyb{w}}- P_{\hat{B}} B^*\hat{\bsyb{w}}^*) \right\|^2 \\
   =& \mathbb{E}_{x\sim \mu} \left\| x^{\top} (\hat{B} \hat{\bsyb{w}}- \hat{B}^*\hat{\bsyb{w}}^*) + x^{\top} P_{\hat{B}}^{\perp} \hat{B}^*\hat{\bsyb{w}}^* \right\|^2 \\
  \leq& 2\times (\mathbb{E}_{x\sim \mu} \left\| x^{\top} (\hat{B} \hat{\bsyb{w}}- \hat{B}^*\hat{\bsyb{w}}^*) \right\|^2 + \left\| x^{\top} P_{\hat{B}}^{\perp} \hat{B}^*\hat{\bsyb{w}}^* \right\|^2) \\
   \leq& 2(\zeta_1 + \zeta_2 \| \hat{\bsyb{w}}^* \|^2) \tag{Definition of $\zeta_1$ and (a)}
\end{align*}

Denote $\Theta = \hat{\bsyb{w}} - P \hat{\bsyb{w}}^*$, also take the spectral decomposition of $\Sigma_t =\sum_{i=1}^{d} \lambda_i \bsyb{v}_i \bsyb{v}_i^{\top}$, here $\langle \bsyb{v}_i, \bsyb{v}_j \rangle=\mathbb{I}[i=j]$. Expand (b) and get
\begin{align*}
    \mathbb{E}_{x\sim \mu} [\Theta^{\top} \hat{B}^{\top} x x^{\top} \hat{B} \Theta] \leq 2(\zeta_1 + \zeta_2 \| \hat{\bsyb{w}}^* \|^2)
\end{align*}
Plug in the definition of $\Sigma_t$ and then use spectral decomposition, we know
\begin{align*}
    \Theta^{\top} \Sigma_t \Theta &\leq 2(\zeta_1 + \zeta_2 \| \hat{\bsyb{w}}^* \|^2) \\
    \sum_{i=1}^d \lambda_i \left\|\bsyb{v}_i^{\top} \Theta \right\|^2 &\leq 2(\zeta_1 + \zeta_2 \| \hat{\bsyb{w}}^* \|^2) 
\end{align*}
Hence we know 
$$\sum_{i=1}^d \left\| \Theta^{\top} \bsyb{v}_i \right\|^2 \leq 2\lambda_{\min}^{-1}(\Sigma_t)(\zeta_1 + \zeta_2 \| \hat{\bsyb{w}}^* \|^2).$$

For any $x$, since $\hat{B}^{\top} x\in \operatorname{span}(\Sigma_t)$, we can write $\hat{B}^{\top} x=\sum_{i=1}^d c_i \bsyb{v}_i$. Then we have the bound for the first term in $(\star)$
\begin{align*}
     \left\| x^{\top} P_{\hat{B}} \hat{B} \Theta \right\|^2
    =&\Theta^{\top} \hat{B}^{\top} x x^{\top} \hat{B} \Theta \\
    =&\sum_{i=1}^d c_i^2 \Theta^{\top} \bsyb{v}_i \bsyb{v}_i^{\top} \Theta \\
    =&\sum_{i=1}^{d} c_i^2 \left\| \Theta^{\top} \bsyb{v}_i \right\|^2\\
    \leq & \sum_{i=1}^d c_i^2 \sum_{i=1}^d \left\| \Theta^{\top} \bsyb{v}_i \right\|^2 \\
    \leq & k^2 \sum_{i=1}^d \left\| \Theta^{\top} \bsyb{v}_i \right\|^2 \tag{$\sum_{i=1}^d c_i^2 = x^{\top} \hat{B}\hat{B}^{\top} x\leq \|x\|^2\cdot \|\hat{B}^{\top}\|^2 \leq k^2$} \\
    \leq & 2k^2(\zeta_1 + \zeta_2 \| \hat{\bsyb{w}}^* \|^2) 
\end{align*}
Plug in the bounds for two terms in $(\star)$ and we get for any $\| x \|\leq M$
\begin{align*}
    \|x^{\top} (\hat{B} \hat{\bsyb{w}} - \hat{B}^* \hat{\bsyb{w}}^*)\| \leq & k \sqrt{2\lambda_{\min}^{-1}(\Sigma_t)(\zeta_1 + \zeta_2 \| \hat{\bsyb{w}}^* \|^2 )} +  \| \hat{\bsyb{w}}^* \|^2 \sqrt{\zeta_2} \\
    \lesssim &   \sqrt{\lambda_{\min}^{-1}(\Sigma_t)\zeta_1} + \| \hat{\bsyb{w}}^* \| \sqrt{\zeta_2 }.
\end{align*}
and thus complete the proof.
\qed 
\\

\subsection{Proof of Theorem 1.} 
\label{sec:thm1}
Now we are ready for proving our main result. Suppose at each level $h$ the algorithm learns representation function $\hat{B}_h^{\top}x$ and weight vector $\bsyb{w}_h$, while the pseudo-ground truth target parameter is $\hat{B}^*_h$ and $\hat{\bsyb{w}}_h^*$. First give expected error bound for Q-value estimation function $x^{\top} \hat{B}_h\hat{\bsyb{w}}_h$ compared with its learning target $x^{\top} \hat{B}^*_h \hat{\bsyb{w}}^*_h$. Denote the distribution from which the training samples are sampled from as $p_x=\Mcal{D}^t, t\in[N]$ and the covariance matrix $\Sigma=\mathbb{E}_{x\sim p_x}[ x x^{\top} ]$. Denote test covariance as $\Sigma'=\mathbb{E}_{x\sim\Mcal{D}^{(T+1)}} [x x^{\top}]$. Expand the expected excess risk $\operatorname{ER}_h$ as 
\begin{align*}
    \operatorname{ER}_h(\hat{B}_h, \hat{\bsyb{w}}_h) = &\frac{1}{2}\mathbb{E}_{x\sim\Mcal{D}^{(T+1)}}[x^{\top} (\hat{B}_{h} \hat{\bsyb{w}}_{T+1,h} - \hat{B}^*_{h} \hat{\bsyb{w}}^*_{T+1,h})]^2\\
    = &\frac{1}{2}\mathbb{E}_{x\sim\Mcal{D}^{(T+1)}}[(\hat{B}_{h} \hat{\bsyb{w}}_{T+1,h} - \hat{B}^*_{h} \hat{\bsyb{w}}^*_{T+1,h})^{\top} x x^{\top} (\hat{B}_{h} \hat{\bsyb{w}}_{T+1,h} - \hat{B}^*_{h} \hat{\bsyb{w}}^*_{T+1,h})] \\
    = & \frac{1}{2} (\hat{B}_{h} \hat{\bsyb{w}}_{T+1,h} - \hat{B}^*_{h} \hat{\bsyb{w}}^*_{T+1,h})^{\top} \Sigma' (\hat{B}_{h} \hat{\bsyb{w}}_{T+1,h} - \hat{B}^*_{h} \hat{\bsyb{w}}^*_{T+1,h}) 
\end{align*}
Apply technical lemma B.2 with $B=[\hat{B}_{h},\hat{B}^*_{T+1,h}]$, we have 
\begin{align*}
    0.9B^{\top}\Sigma' B\preceq \frac{1}{n} B^{\top}X^{\top}_{T+1}X_{T+1}B
\end{align*}
while implies $0.9\bsyb{v}^{\top}B^{\top}\Sigma' B \bsyb{v}\leq \frac{1}{n}\bsyb{v}^{\top}B^{\top}X^{\top}_{T+1}X_{T+1}B\bsyb{v}$ for $\bsyb{v}=\left[\begin{matrix}\hat{\bsyb{w}}_{T+1,h}\\ -\hat{\bsyb{w}}^*_{T+1,h} \end{matrix}\right]$. And this inequality becomes
\begin{align}
    &(\hat{B}_h \hat{\bsyb{w}}_{T+1,h} - B^*_h \hat{\bsyb{w}}^*_{T+1,h})^{\top} \Sigma' (\hat{B}_h \hat{\bsyb{w}}_{T+1,h} - \hat{B}^*_h\hat{\bsyb{w}}^*_{T+1,h})\\
    \leq&\frac{1}{0.9n_2} (\hat{B}_h \hat{\bsyb{w}}_{T+1,h} - \hat{B}^*_h \hat{\bsyb{w}}^*_{T+1,h})^{\top} X_{T+1}^{\top} X_{T+1} (\hat{B}_h \hat{\bsyb{w}}_{T+1,h} - \hat{B}^*_h\hat{\bsyb{w}}^*_{T+1,h})
\end{align}

Hence we have 
\begin{align*}
    \operatorname{ER}_h(\hat{B}_h, \hat{\bsyb{w}}_{T+1,h})\leq& \frac{1}{1.8n_2}(\hat{B}_h \hat{\bsyb{w}}_{T+1,h} - \hat{B}^*_h \hat{\bsyb{w}}^*_{T+1,h})^{\top} X^{\top}_{T+1} X_{T+1} ( \hat{B}_h \hat{\bsyb{w}}_{T+1,h} - \hat{B}^*_h \hat{\bsyb{w}}^*_{T+1,h})\\
    =&\frac{1}{1.8n_2}\|X_{T+1}(\hat{B}_h \hat{\bsyb{w}}_{T+1,h} - \hat{B}^*_h\hat{\bsyb{w}}^*_{T+1,h})\|^2
\end{align*}

The optimality of $\hat{\bsyb{w}}_{T+1,h}$ implies that 
\begin{align*}
    X_{T+1}\hat{B}_h \hat{\bsyb{w}}_{T+1,h}&=P_{X_{T+1}\hat{B}}(y_{T+1})\\
    &=P_{X_{T+1}\hat{B}}(X_{T+1}B^*_h\hat{\bsyb{w}}^*_{T+1,h}+z_{T+1,h})
\end{align*}

By the calculation above, it follows the upper bound for excess risk at level $h$ will not exceed
\label{equ:eq7}
\begin{align}
    \operatorname{ER}_h(\hat{B}_h,\hat{\bsyb{w}}_{T+1,h}) &\lesssim \frac{1}{n} \left\| P_{X_{T+1} \hat{B}_h } (X_{T+1} \hat{B}^*_h \hat{\bsyb{w}}^*_{T+1,h} + z_{T+1,h}) - X_{T+1} \hat{B}^*_h \hat{\bsyb{w}}^*_{T+1,h}\right\|^2\\
    &=\frac{1}{n}\left\|-P_{X_{T+1}\hat{B}_h}^{\perp} X_{T+1} \hat{B}^*_h \hat{\bsyb{w}}^*_{T+1,h} + P_{X_{T+1},\hat{B}_h} z_{T+1,h} \right\|^2\\
    &=\frac{1}{n}\left\|P_{X_{T+1}\hat{B}_h}^{\perp} X_{T+1}\hat{B}^*_h \hat{\bsyb{w}}^*_{T+1,h} \right\|^2_F + \frac{1}{n} \left\|P_{X_{T+1},\hat{B}_h} z_{T+1,h}\right\|^2_F
\end{align}

For the first term of (\hyperref[equ:eq7]{7}), when the distribution from which $X_{T+1}$ sampled is constrained within space $\hat{B}_h$, it simply becomes zero. Otherwise, we can still use Lemma A.2 to give it a bound, which is the same as following inequality (\hyperref[equ:9]{8}). Applying technical lemma B.3, we know the second term has an upper bound with probability at least $1-\frac{\delta}{5H}$,  
\label{equ:bnd1}
\begin{align*}
 \left\|P_{X_{T+1}\hat{B}_h} z_{T+1,h} \right\|_F^2 \lesssim (\sigma+1)^2 \left( d+\log\frac{H}{\delta}\right).   \tag{Bound 1}
\end{align*}

Next, we are going to bound the difference between space $\operatorname{span}(\hat{B}_h)$ and $\operatorname{span}(\hat{B}_h^*)$. According to lemma A.2, we know with probability $1-O(\delta/H)$
\label{equ:9}
\begin{align}
    \left\|P_{\Sigma^{1 / 2}\hat{B}}^{\perp} \Sigma^{1 / 2} \hat{B}^{*}_h \right\|_{F}^{2}
    \lesssim\frac{(\sigma+1)^2(Td+Dd\log( N)+\log\frac{H}{\delta})}{N\sigma_d^2(\hat{W}^*_h)}
\end{align}
An important issue here is that, $\hat{W}_h^*=[\hat{\bsyb{w}}^*_{1,h+1},\hdots,\hat{\bsyb{w}}^*_{1,h+1}]$ is the induced by equation
\begin{align*}
    \hat{\bsyb{w}}^*_{t,h+1}&=\theta+\sum_{s'} \psi(s')\cdot \hat{V}_{h+1}^{t} (s')\\
    \hat{V}_{h+1}^t (s') &= \max_{a} \xi(s,a)^{\top} \hat{B}_{h+1} \hat{\bsyb{w}}_{t,h+1},
\end{align*}
therefore we cannot directly apply assumption 2. However we can utilize lemma A.3 to estimate their difference.  When $N\gg D^2 d^2 \log(N)$, we have $|\sigma_d^2(W^*_h) - \sigma_d^2(\hat{W}^*_h)| \leq \sum_{t=1}^T \|\hat{\bsyb{w}}_{t,h} -\hat{\bsyb{w}}^*_{t,h} \|^2 = o\left(\frac{T(H-h+1)^2}{d^2}\right)$. Thus we know $\sigma_d^2(\hat{W}^*_h)\geq \sigma_d^2(W^*_h)+ o\left(\frac{T(H-h+1)^2}{d}\right) =\Omega((H-h+1)^2 T/d)$. Plug it into (8) and we get
\begin{align*}
    \left\|P_{\Sigma^{1 / 2}\hat{B}}^{\perp} \Sigma^{1 / 2} \hat{B}^{*}_h \right\|_{F}^{2}
    \lesssim &\frac{(\sigma+1)^2(Td^2+Dd^2\log( N)+d\log\frac{H}{\delta})}{N T(H-h+1)^2}\\
    \leq& \frac{(\sigma+1)^2}{(H-h+1)^2}\left(\frac{d^2+\log \frac{H}{\delta}}{N}+\frac{Dd^2\log( N)}{NT}\right) \tag{$T>d$}
\end{align*}
By learning algorithm we know $\Sigma=I_D/D$, therefore $\Sigma^{1/2}=I_D/\sqrt{D}$. Plug into inequality above to get
\label{equ:bnd2}
\begin{align*}
    \left\|P_{\hat{B}}^{\perp} \hat{B}^{*}_h \right\|_{F}^{2} \lesssim  \frac{(\sigma+1)^2D}{(H-h+1)^2}\left(\frac{d^2+\log \frac{H}{\delta}}{N}+\frac{Dd^2\log( N)}{N T}\right). \tag{Bound 2}
\end{align*}
We can then combine \hyperref[equ:bnd1]{Bound 1} and \hyperref[equ:bnd2]{Bound 2} by plugging them into key lemma. Denote $\kappa:=\lambda_{\min}^{-1}(\mathbb{E}_{x\sim \Mcal{D}^{(T+1)}}[\hat{B}^{\top}_h x x^{\top} \hat{B}_h])$, we know for any $x$ bounded by $\|x\|\leq 1$, we have with probability at least $1-O(\delta/H)$,
\begin{align}
    &\left\| x^{\top}(\hat{B}_{T+1,h} \hat{\bsyb{w}}_{T+1,h} - \hat{B}^*_{T+1,h} \hat{\bsyb{w}}^*_{T+1,h}) \right\| \\
    \lesssim& (\sigma+1)\left[ \sqrt{\frac{ \kappa (d+\log \frac{H}{\delta}) }{n}} + \frac{\|\hat{\bsyb{w}}_{T+1,h}^* \|}{H - h + 1}\sqrt{ \frac{D(d^2+\log \frac{H}{\delta})}{N}+\frac{D^2d^2\log( N)}{N T} }\ \right]
\end{align}
 From lemma A.3 we know because $n\gg d^2$, $\|\hat{\bsyb{w}}_{T+1,h}^* \| - \|\bsyb{w}_{T+1,h}^* \|=o(H-h+1)$, hence $\|\hat{\bsyb{w}}_{T+1,h}^* \|=\Theta(H-h+1)$. Therefore we get the Q-value prediction error compared with pseudo-ground truth as 
\begin{align*}
    &\left\| x^{\top}(\hat{B}_{T+1,h} \hat{\bsyb{w}}_{T+1,h} - \hat{B}^*_{T+1,h} \hat{\bsyb{w}}^*_{T+1,h}) \right\| \\
    \lesssim& (\sigma+1) \left[ \sqrt{\frac{ \kappa(d+\log \frac{H}{\delta}) }{n}} + \sqrt{\frac{D(d^2+\log \frac{H}{\delta})}{N}+\frac{D^2d^2\log(N)}{N T}}\ \right] \\
    \lesssim& (\sigma+1)  \sqrt{\frac{ \kappa(d+\log \frac{H}{\delta}) }{n} + \frac{D(d^2+\log \frac{H}{\delta})}{N}+\frac{D^2d^2\log(N)}{N T}}
\end{align*}

Finally, we bridge between Q-value estimation error and policy's sub-optimality. For the sake of simplicity, we use $\|x^{\top}(\hat{B}_h \hat{\bsyb{w}}_h - \hat{B}_h^* \hat{\bsyb{w}}_h^*)\| \leq \epsilon_h$ to denote the prediction error at level $h$. Denote the policy induced by $\hat{B}_h\hat{\bsyb{w}}_h$ as $\pi_h(\cdot)=\arg\max_{a} \xi(\cdot,a)^{\top} \hat{B}_h\hat{\bsyb{w}}_h$, and define corresponding value function $V^{\pi}_h(s)$ as
\begin{align*}
    V^{\pi}_h(s):=\mathbb{E}\left[ \sum_{k=h}^H r_{k}(s_h,\pi(s_h))\Big| s_h=s \right],
\end{align*}
and optimal value function as 
\begin{align*}
    V^{\star}_h(s):=&\mathbb{E}\left[ \sum_{k=h}^H r_{k}(s_h,\pi^{\star}(s_h))\Big| s_h=s \right]\\
    =&\max_a \xi(s,a)^{\top} B^* \bsyb{w}^*_h.
\end{align*}
The estimation value for each state at level $h$ by $\hat{B}_h\hat{\bsyb{w}}_h$ and $\hat{B}^*_h \hat{\bsyb{w}}^*_h$ are
\begin{align*}
    \hat{V}^{\pi}_h(s)&:=\max_{a} \xi(s,a)^{\top} \hat{B}_h \hat{\bsyb{w}}_h,\\
    \tilde{V}^{\pi}_h(s)&:=\max_{a} \xi(s,a)^{\top} \hat{B}^*_h \hat{\bsyb{w}}^*_h.
\end{align*}
Setting anchor value as optimal value function $V^{\star}(\cdot)$ in lemma A.4, we know our estimation of value for each state $s$ as $\hat{V}_h (s) = \max_{a}\xi(s,a)^{\top} \hat{B}_h \hat{\bsyb{w}}_h$ will not exceed the summation of Q-value prediction error for all following levels, namely 
\begin{align*}
    &\left| V^{\star}_h(s) - \hat{V}_h(s) \right| \leq  \sum_{k=h}^H \epsilon_k\\
    =&(\sigma+1) (H-h+1) \sqrt{\frac{ \kappa(d+\log \frac{H}{\delta}) }{n} + \frac{D(d^2+\log \frac{H}{\delta})}{N}+\frac{D^2d^2\log(N)}{N T}}
\end{align*}

After establishing the difference between estimation value and optimal value, we again recursively use math induction from $H$ to 1 to show that $\left| V^{\pi}_h (s) - \hat{V}^{\pi}_h (s) \right| \leq \sum_{k=h}^{H} \epsilon_k, \text{ for } \forall s \in \Mcal{S}$.

Notice this is trivial for $h=H+1$ since the value for all states are 0. Suppose the claim holds for $h+1$, then by definition we have
\begin{align*}
    &\left| V^{\pi}_h (s) - \hat{V}^{\pi}_h (s) \right| \\
  =&\left| V^{\pi}_h (s) - \tilde{V}^{\pi}_h(s) + \tilde{V}^{\pi}_h(s) - \hat{V}^{\pi}_h (s) \right| \\
 \leq& \underbrace{\left| V^{\pi}_h (s) - \tilde{V}^{\pi}_h(s) \right|}_{q_1} + \underbrace{\left| \tilde{V}^{\pi}_h(s) - \hat{V}^{\pi}_h (s) \right|}_{q_2}\\
\end{align*}
 First focus on $q_2$. Denote $a_1=\arg\max_{a} \xi(s,a)^{\top} \hat{B}_h \hat{\bsyb{w}}_h$ and $a_2=\arg\max_{a} \xi(s,a)^{\top} \hat{B}^*_h \hat{\bsyb{w}}^*_h$. if $\xi(s,a_1)^{\top} \hat{B}_h \hat{\bsyb{w}}_h-\xi(s,a_2)^{\top} \hat{B}^*_h \hat{\bsyb{w}}^*_h > 0$ then according to definition and the optimality of $a_2$ for $\hat{B}^*_h\bsyb{w}^*_h$ we have
\begin{align*}
    q_2 =&\xi(s,a_1)^{\top} \hat{B}_h \hat{\bsyb{w}}_h-\xi(s,a_2)^{\top} \hat{B}^*_h \hat{\bsyb{w}}^*_h\\
    \leq &\xi(s,a_1)^{\top} \hat{B}_h \hat{\bsyb{w}}_h - \xi(s,a_1)^{\top} \hat{B}^*_h \hat{\bsyb{w}}^*_h \\
       + &\xi(s,a_1)^{\top} \hat{B}^*_h \hat{\bsyb{w}}^*_h - \xi(s,a_2)^{\top} \hat{B}^*_h \hat{\bsyb{w}}^*_h \\
       & \tag{By $a_2$'s optimality, $<0$} \\
    \leq & \xi(s,a_1)^{\top} (\hat{B}_h \hat{\bsyb{w}}_h - \hat{B}^*_h \hat{\bsyb{w}^*}_h)\\
    \leq & \epsilon_h \tag{error bound for level $h$}
\end{align*}
Similar results holds for $\xi(s,a_1)^{\top} \hat{B}_h \hat{\bsyb{w}}_h-\xi(s,a_2)^{\top} \hat{B}^*_h \hat{\bsyb{w}}^*_h<0$,
\begin{align*}
    q_2 =&\xi(s,a_2)^{\top} \hat{B}^*_h \hat{\bsyb{w}}^*_h-\xi(s,a_1)^{\top} \hat{B}_h \hat{\bsyb{w}}_h\\
    \leq &\xi(s,a_2)^{\top} \hat{B}^*_h \hat{\bsyb{w}}^*_h - \xi(s,a_2)^{\top} \hat{B}_h \hat{\bsyb{w}}_h \\
       + &\xi(s,a_2)^{\top} \hat{B}_h \hat{\bsyb{w}}_h - \xi(s,a_1)^{\top} \hat{B}_h \hat{\bsyb{w}}_h\\
       & \tag{By $a_1$'s optimality, $<0$}\\
    \leq & \xi(s,a_2)^{\top} (\hat{B}^*_h \hat{\bsyb{w}}^*_h - \hat{B}_h \hat{\bsyb{w}^*}_h)\\
    \leq & \epsilon_h \tag{error bound for level $h$}
\end{align*}
 
 Next, without the loss of generality we assume $ V^{\pi}_h (s) - \tilde{V}^{\pi}_h(s)>0$. For any $s$, denote $a=\pi(s)$ and $\hat{a}=\arg\max_a \xi(s,a)^{\top} \hat{B}^*_h \hat{\bsyb{w}}^*_h$, then we get
\begin{align*}
    q_1 =& V^{\pi}_h (s) - \tilde{V}^{\pi}_h(s)\\
    =& r_h(s,a) + [\Mcal{T}_{a}V^{\pi}_{h+1}](s) - \xi(s, \hat{a})^{\top} \hat{B}^*_h \hat{\bsyb{w}}^*_h\\
    \leq& r_h(s,a) + [\Mcal{T}_{a}V^{\pi}_{h+1}](s) - \xi(s, a)^{\top} \hat{B}^*_h \hat{\bsyb{w}}^*_h\\
    +&\xi(s, a)^{\top} \hat{B}^*_h \hat{\bsyb{w}}^*_h - \xi(s, \hat{a})^{\top} \hat{B}^*_h \hat{\bsyb{w}}^*_h \tag{$<0$}\\
    \leq & r_h(s,a) + [\Mcal{T}_{a}V^{\pi}_{h+1}](s) - \xi(s, a)^{\top} \hat{B}^*_h \hat{\bsyb{w}}^*_h\\
    =& [\Mcal{T}_{a} (V^{\pi}_{h+1}-\hat{V}^{\pi}_{h+1})](s)\\
    \leq & \max_{s} \left| V^{\pi}_{h+1}(s)-\hat{V}^{\pi}_{h+1}(s) \right| \tag{$*$}
\end{align*}
By induction we know $q_1\leq (*)\leq \sum_{k=h+1}^{H}\epsilon_k$, hence we can take summation of $q_1$ and $q_2$ and conclude that our claim holds for any $h\in[H]$. So eventually, take union bound over all level $h\in [H]$, we have sub-optimality bound holds with probability at least $1-\delta$.
\begin{align*}
    &\left| V^{\star}_h(s) - {V}^{\pi}_h(s) \right| \leq | V^{\star}_h(s) - {V}^{\pi}_h(s) | + | V^{\pi}_h(s) - \hat{V}_h(s) |=2\sum_{k=h}^H \epsilon_k\\
    \lesssim &(\sigma+1) (H-h+1)\sqrt{\frac{ \kappa(d+\log \frac{H}{\delta}) }{n} + \frac{D(d^2+\log \frac{H}{\delta})}{N}+\frac{D^2d^2\log(N)}{N T}}
\end{align*}

Accounting that each sample essentially contains at most $m\leq H^2$ tuples $(s,a,r(s,a),s')$, the authentic bound should divide each $N$ and $n$ by $H^2$. Therefore the sub-optimality at first level satisfies following bound with probability at least $1-\delta$

\begin{align*}
    \left| V^{\star}(s) - {V}^{\pi}(s) \right|
    \lesssim &(\sigma+1) H^2\sqrt{\frac{ \kappa (d+\log \frac{H}{\delta}) }{n} + \frac{D(d^2+\log \frac{H}{\delta})}{N}+\frac{D^2d^2\log(N)}{N T}}
\end{align*}
\qed
\\

\subsection{Technical Lemmas}
Used technical lemmas' proof are listed below. Since we use 1-sub-Gaussian distribution $\Mcal{D}^t$, all following lemma involving $\rho$ can be applied with $\rho=1$. \\

\textbf{Technical Lemma B.0} (Covariance Concentration Bound) \textit{Let $\bsyb{a}_1, \bsyb{a}_2,\hdots,\bsyb{a}_n$ be i.i.d. d-dimensional random vectors such that $\mathbb{E}[\bsyb{a}_i]=0,\mathbb{E}[\bsyb{a}_i\bsyb{a}_i^{\top}]=I$ and each $\bsyb{a}_i$ is $\rho^2$ sub-Gaussian. For $\delta\in(0,1)$, if $N\gg \rho^4(d+\log(1/\delta))$, then with probability $1-\delta$ we have}
\begin{align*}
    0.9I \preceq \frac{1}{N}\sum_{i=1}^N \bsyb{a}_i\bsyb{a}_i^{\top} \preceq 1.1I
\end{align*}
\textit{Proof.}
Let $A=\frac{1}{N}\sum_{i=1}^{N}\bsyb{a}_i\bsyb{a}_i^{\top}-I$. It suffice to show that $\|A\| \leq 0.1$ with probability $1-\delta$. We will prove by $\epsilon$-net argument for $\Mcal{S}^{d-1}=\{v:v^{\top}v=1\}$. For any fixed $v\in\Mcal{S}^{d-1}$, $v^{\top}Av=\frac{1}{N}\sum_{i=1}^N[(v^{\top}\bsyb{a}_i)^2-1]$. From assumption we know that $v^{\top}\bsyb{a}_i$ has mean 0, variance 1 and is $\rho^2$ sub-Gaussian (notice that $\rho\geq 1$). Therefore $(v^{\top}\bsyb{a}_i)^2-1$ is zero-mean and $16rho^2$-sub-exponential. By Bernstein inequality for sub-exponential random variables, we have for any $\epsilon>0$
\begin{align*}
    Pr\left[|v^{\top}Av|>\epsilon\right]\leq 2\exp\left(-\frac{N}{2}\min\left\{ \frac{\epsilon^2}{(16\rho^2)^2},\frac{\epsilon}{16\rho^2}\right\}\right).
\end{align*}

Take a $\frac{1}{5}$-net $\Mcal{N}\subset \Mcal{S}^{d-1}$ with size $|\Mcal{N}  |\leq e^{O(d)}$. By union bound over all $v\in\Mcal{N}$, we have
\begin{align*}
    Pr\left[\max_{v\in\Mcal{N}}|v^{\top}Av|>\epsilon\right]\leq& 2|\Mcal{N}|\exp\left(-\frac{N}{2}\min\left\{ \frac{\epsilon^2}{(16\rho^2)^2},\frac{\epsilon}{16\rho^2}\right\}\right)\\
    \leq& 2\exp\left(O(d)-\frac{N}{2}\min\left\{ \frac{\epsilon^2}{(16\rho^2)^2},\frac{\epsilon}{16\rho^2}\right\}\right).
\end{align*}
Plug in $\epsilon=\frac{1}{20}$ and notice that $\rho>1$, inequality above becomes
\begin{align*}
    Pr\left[\max_{v\in\Mcal{N}}|v^{\top}Av|>\frac{1}{20}\right]\leq\exp\left(O(d)-\frac{N}{2}\cdot\frac{(1/20)^2}{(16\rho^2)^2}\right)\leq \delta,
\end{align*}
And the last inequality is equal to $N\gg \rho^4(d+\log(1/\delta))$ hence it holds.

Therefore we know with probability at least $1-\delta$, $[\max_{v\in\Mcal{N}}|v^{\top}Av|\leq\frac{1}{20}$. And notice that by the definition of $1/5$-net, for arbitrary $u\in\Mcal{S}^{d-1}$, there exists $u'\in\Mcal{N}$ such that $\|u-u'\|\leq \frac{1}{5}$. Then we have 
\begin{align*}
    \left\|u^{\top}Au\right\|\leq&\left\|(u')^{\top}Au'\right\|+2\left\|(u-u')^{\top}Au'\right\|+\left\|(u-u')^{\top}A(u-u')\right\|\\
    \leq&\frac{1}{20}+2\|u-u'\|\cdot\|A\|\cdot\|u'\|+\|u-u'\|^2\cdot\|A\|\\
    \leq&\frac{1}{20}+\frac{2}{5}\|A\|+\|A\|\cdot\frac{1}{5^2}\\
    \leq & \frac{1}{20}+\frac{1}{2}\|A\|.
\end{align*}
Taking a supreme over $u\in\Mcal{S}^{d-1}$ we obtain $\|A\|\leq \frac{1}{20}+\frac{1}{2}\|A\|$, which implies $\|A\|\leq\frac{1}{10}$ and complete the proof.
\qed
\\

\textbf{Technical Lemma B.1} (Covariance Concentration of Source Tasks) \textit{Suppose $N\gg \rho^4(D+\log(T/\delta))$ for $\delta\in(0,1)$. Then with probability at least $1-\frac{\delta}{10}$ over the $X_1,\hdots,X_T$ in the source tasks, we have}
\begin{align}
    0.9\Sigma\preceq \frac{1}{N}X_t^{\top}X_t\preceq 1.1\Sigma,\quad \forall t\in[T].
\end{align}
\textit{Proof}. According to assumption on $\bar{p}_x$, we can denote $X_{t}=\Sigma^{1/2}\bar{X}_{t}$ where $\bar{X}_{t}\in\mathbb{R}^{N\times D}$ are the rows of $\bar{X}_t$ holds i.i.d. samples of $\bar{p}_x$. Since they satisfy the condition in Lemma B.0, we know that with probability at least $1-\frac{\delta}{10T}$ 
\begin{align*}
    0.9I\preceq \frac{1}{N}\bar{X}_t\bar{X}_t^{\top}\preceq 1.1I,t\in[T]
\end{align*}
which implies 
$$0.9\Sigma\preceq \frac{1}{N}\Sigma^{1/2}\bar{X}_t^{\top}\bar{X}_t\Sigma^{1/2}=\frac{1}{N}X_t^{\top}X_t\preceq 1.1\Sigma.$$
Finally take a union bound over all $T$ tasks and finish the proof.
\qed
\\

\textbf{Technical Lemma B.2} (Covariance Concentration of Target Task). \textit{Suppose $n\gg \rho^4(d+\log(1/\delta))$ for $\delta\in(0,1)$. Then for any given matrix $B\in\mathbb{R}^{D\times 2d}$ that is independent of $X_{T+1}$, with probability at least $1-\frac{\delta}{10}$ over $X_{T+1}$ we have}
\begin{align}
    0.9B^{\top}\Sigma B \preceq \frac{1}{n}B^{\top}X_{T+1}^{\top}X_{T+1}B\preceq 1.1 B^{\top}\Sigma B.
\end{align}
\textit{Proof}. Similarly, denote $X_{T+1}=\Sigma^{1/2}\bar{X}_{T+1}$ where $\bar{X}_{T+1}\in\mathbb{R}^{n\times D}$ are the rows of $\bar{X}_t$ holds i.i.d. samples of $\bar{p}_x$. Take the SVD of $\Sigma^{1/2}B$ as $\Sigma^{1/2}B=UDV^{\top}$, where $U\in \mathbb{R}^{D\times 2d}$ has orthonormal columns. And $\bar{X}_{T+1}U\in\mathbb{R}^{n\times 2d}$ are i.i.d. 2d-dimensional random vectors with zero mean, identity covariance and $\rho^2$-subGaussian. Therefore we can apply Lemma B.0 and get with probability at least $1-\delta/10$
\begin{align*}
    0.9I\preceq U^{\top}\bar{X}_{T+1}^{\top}\bar{X}_{T+1}U\preceq 1.1 I
\end{align*}
Multiply $VD$ to left side and $DV^{\top}$ to the right, we get
\begin{align*}
    0.9 V D D V^{\top} \preceq \frac{1}{n} V D U^{\top} \bar{X}_{T+1}^{\top} \bar{X}_{T+1} U D V^{\top} \preceq 1.1 V D D V^{\top}
\end{align*}
Plug in $VDDV^{\top}=VDU^{\top}UDV^{\top}=B^{\top}\Sigma B$ and $UDV^{\top}=\Sigma^{1/2}B$, then we know the conclusion holds.
\qed
\\

\textbf{Technical Lemma B.3} (Concentration guarantee of noise) \textit{Under the setting of the noise at level $h$ as $z_h=z_h^R+z_h^T$, suppose there are $N$ i.i.d. random variable $z_h^{(j)},j\in[N]$ drawn from the distribution of $z_h$, or compactly written as $Z\in\mathbb{R}^N$. We have the following bound holds with probability at least $1-O(\delta)$}
\begin{align*}
    \|Z\|^2=\sum_{j=1}^N \|z_h^{j}\|^2\lesssim (\sigma+1)^2(N+\log(1/\delta))
\end{align*}

\textit{Proof.} Notice that by induction of $h$ from $H$ to $1$, we know $Q_h(s,a) \in [0,(H-h+1)(1+\sigma)]$, thus  $V_h(s)=\max_{a\in\Mcal{A}}Q_h(s,a)\in[0,(H-h+1)(1+\sigma)]$ for $\forall s\in\Mcal{S},a\in\Mcal{A}$. So we can factor each $z_h^{(j)}$ into the average of $m=(H-h+1)^2$ i.i.d. random variables $X^{(j)}_i=V_{h+1}^{(t)}(s_{jk}')-\mathbb{E}_{s'\sim P_{s_j^t,a_j^t}}V^{(t)}_{h+1}(s')$ as $$z_h^{(j)}=\frac{1}{(H-h+1)^2}\sum_{i=1}^{(H-h+1)^2} X^{(j)}_i$$
And each $\|X^{(j)}_{i}\|$ is bounded by $2(H-h)(1+\sigma)$ because
\begin{align*}
    \left\|X^{(j)}_i\right\|\leq&\|V^{(t)}_{h+1}(s'_{jk})\|+\left\|\mathbb{E}_{s'\sim P_{s_j^t,a_j^t}}V^{(t)}_{h+1}(s')\right\|\\
    \leq &(H-h)(1+\sigma)+(H-h)(1+\sigma)\\
    =&2(H-h)(1+\sigma)
\end{align*}

Hence we know each $X_i^{(j)}$ is $4(H-h)^2(1+\sigma)^2$ Sub-Gasussian, therefore $z_h^T$ is sub-Gaussian with parameter $\sigma_T^2=\sum_{j=1}^{(H-h+1)^2} \frac{4(H-h)^2(1+\sigma)^2}{(H-h+1)^4}\leq 4(1+\sigma)^2$. Combined with $z_h^R$ being naturally sub-Gasussian with parameter $\sigma$, we know $z_h=z_h^T+z_R^T$ is a sub-Gaussian variable with parameter $\sigma^2+4(1+\sigma)^2\leq 5(\sigma+1)^2$. By technical lemma B.5 we know $z_h^{(j)}$ satisfy
$$\mathbb{E}[e^{\lambda(\|z_h^{(j)}\|^2-5(\sigma+1)^2)}]\leq \exp\left(\frac{100 \lambda^2 (1+\sigma)^4}{2}\right),\quad 0\leq \lambda< 1/20(1+\sigma)^2$$

Hence we know $\|Z\|^2-5N(\sigma+1)^2=\sum_{j=1}^N( \|z_h^{(j)}\|^2-5(\sigma+1)^2)$ is sub-exponential with parameter $\nu_*=2\cdot 5(1+\sigma)^2\sqrt{N}=10(1+\sigma)^2\sqrt{N},\alpha_*=4\cdot 5(1+\sigma)^2=20(1+\sigma)^2$. Plug in the bound in technical lemma B.4, we know when $t>\frac{\nu^{2}_*}{\alpha_*^2}=5N(1+\sigma)^2$ the tail probability bound for $\|Z\|^2$ is
\begin{align*}
    P(\|Z\|^2-9N(\sigma+1)^2>t)&\leq e^{-\frac{t}{40(1+\sigma)^2}}
\end{align*}
let $\delta=e^{-\frac{t}{40(1+\sigma)^2}}$, then we know with probability at least $1-\delta$
\begin{align*}
    \|Z\|^2\leq& 5N(\sigma+1)^2+t\\
    =&9N(\sigma+1)^2+40(\sigma+1)^2\log(1/\delta)\\
    \lesssim&(\sigma+1)^2(N+\log(1/\delta))
\end{align*}

\textbf{Technical Lemma B.4} \textit{If two matrices $A_1$ and $A_2$ have same number of columns and satisfy $A_1^{\top}A_1\succeq A_2^{\top}A_2$, then for any matrix B we have}
$$A_1^{\top}P_{A_1B}^{\perp}A_1\succeq A_2^{\top}P_{A_2B}^{\perp}A_2$$
Also, it implies for any compatible sized matirx $C$
$$\left\|P_{A_1B}^{\perp}A_1C\right\|_F^2\geq \left\|P_{A_2B}^{\perp}A_2C\right\|_F^2$$
\textit{Proof.} For the first part, it suffices to show the following holds for any vecotr $\bsyb{v}$
$$
\boldsymbol{v}^{\top} A_{1}^{\top} P_{A_{1} B}^{\perp} A_{1} \boldsymbol{v} \geq \boldsymbol{v}^{\top} A_{2}^{\top} P_{A_{2} B}^{\perp} A_{2} \boldsymbol{v}
$$
which is equivalent to 
$$
\min _{\boldsymbol{w}}\left\|A_{1} B \boldsymbol{w}-A_{1} \boldsymbol{v}\right\|_{2}^{2} \geq \min _{\boldsymbol{w}}\left\|A_{2} B \boldsymbol{w}-A_{2} \boldsymbol{v}\right\|_{2}^{2}
$$
Let $\boldsymbol{w}^{*} \in \arg \min _{\boldsymbol{w}}\left\|A_{1} B \boldsymbol{w}-A_{1} \boldsymbol{v}\right\|_{2}^{2}$, then we have
\begin{align*}
\min _{\boldsymbol{w}}\left\|A_{1} B \boldsymbol{w}-A_{1} \boldsymbol{v}\right\|_{2}^{2} &=\left\|A_{1} B \boldsymbol{w}^{*}-A_{1} \boldsymbol{v}\right\|_{2}^{2} \\
&=\left(B \boldsymbol{w}^{*}-\boldsymbol{v}\right)^{\top} A_{1}^{\top} A_{1}\left(B \boldsymbol{w}^{*}-\boldsymbol{v}\right) \\
& \geq\left(B \boldsymbol{w}^{*}-\boldsymbol{v}\right)^{\top} A_{2}^{\top} A_{2}\left(B \boldsymbol{w}^{*}-\boldsymbol{v}\right) \\
&=\left\|A_{2} B \boldsymbol{w}^{*}-A_{2} \boldsymbol{v}\right\|_{2}^{2}\\
&\geq \min _{w}\left\|A_{2} B \boldsymbol{w}-A_{2} \boldsymbol{v}\right\|_{2}^{2}
\end{align*}
finishing the proof of the first part.\\

For the second part, from $A_{1}^{\top} P_{A_{1}}^{\perp} A_{1} \succeq A_{2}^{\top} P_{A_{2} B}^{\perp} A_{2}$ we know
$$
\left(B^{\prime}\right)^{\top} A_{1}^{\top} P_{A_{1}}^{\perp} A_{1} B^{\prime} \succeq\left(B^{\prime}\right)^{\top} A_{2}^{\top} P_{A_{2} B}^{\perp} A_{2} B^{\prime}
$$
Taking trace on both sides and we get
$$
\left\|P_{A_{1} B}^{\perp} A_{1} B^{\prime}\right\|_{F}^{2} \geq\left\|P_{A_{2} B}^{\perp} A_{2} B^{\prime}\right\|_{F}^{2},
$$
which finishes the whole proof.
\qed 

\textbf{Technical Lemma B.5} \textit{If a random variable $X$ is $\sigma^2$ sub-Gaussian, then random variable $Y=X^2-\sigma^2$ is sub-exponential with parameter $(\nu,\alpha)=\left(2\sigma^2,4\sigma^2\right)$, which says}
$$\mathbb{E}\left[e^{\lambda(Y-\sigma^2)}\right]\leq e^{\frac{\lambda^2\nu^2}{2}},\quad for\ \forall |\lambda|<\frac{1}{\alpha}$$
As a corollary, we have the bound 
$$P(Y>\sigma^2+t)\leq e^{-\frac{t}{2\alpha}},\quad t>\sigma^2$$
\textit{Proof.} We first prove that $\mathbb{E}[e^{\lambda X^2}]\leq \frac{1}{\sqrt{1-2\sigma^2\lambda}}$ for all $\lambda \in[0,1/2\sigma^2)$. By the definition of sub-Gaussian, we know $\mathbb{E}[e^{\lambda X}]\leq e^{\frac{s^2\lambda^2}{2}}$. Multiply both side by $e^{-\frac{\lambda^2\sigma^2}{2s}}$ and we get 
\begin{align*}
    \mathbb{E}\left[e^{\lambda X-\frac{\lambda^{2} \sigma^{2}}{2 s}}\right] \leq e^{\frac{\lambda^{2} \sigma^{2}(s-1)}{2 s}},\quad (*)
\end{align*}
Since the inequality holds for all $\lambda \in \mathbb{R}$, we can integrate over $\lambda$ and use Fubini's theorem to exchanging the order of integration and expectation. \\

On the right handside we 
$$\int_{-\infty}^{\infty} \exp \left(\frac{\lambda^{2} \sigma^{2}(s-1)}{2 s}\right) d \lambda=\frac{1}{\sigma} \sqrt{\frac{2 \pi s}{1-s}}$$
On the left handside, for any fixed $x\in \mathbb{R}$ we first do the integral on $\lambda$ and get
$$\int_{-\infty}^{\infty} \exp \left(\lambda x-\frac{\lambda^{2} \sigma^{2}}{2 s}\right) d \lambda=\frac{\sqrt{2 \pi s}}{\sigma} e^{\frac{s x^{2}}{2 \sigma^{2}}}$$
Take the expectation of last equation over $x$ and from $(*)$ we know
\begin{align*}
    \frac{\sqrt{2\pi s}}{\sigma}\mathbb{E}\left[e^{\frac{sX^2}{2\sigma^2}}\right]&\leq \frac{1}{\sigma}\sqrt{\frac{2\pi s}{1-s}}\\
    \mathbb{E}\left[e^{\frac{sX^2}{2\sigma^2}}\right]&\leq \frac{1}{\sqrt{1-s}}\\
    \mathbb{E}[e^{\lambda X^2}]&\leq \frac{1}{\sqrt{1-2\sigma^2\lambda}}\quad ,\lambda=\frac{s}{2\sigma^2}
\end{align*}
Thus by some simple calculus we know
\begin{align*}
    \mathbb{E}\left[e^{\lambda(Y-\sigma^2)}\right]&\leq \frac{e^{-\lambda \sigma^2}}{\sqrt{1-2\sigma^2\lambda}}\\
    &\leq e^{2\lambda^2\sigma^4},\quad \forall |\lambda| \leq \frac{1}{4\sigma^2}
\end{align*}
Hence we know $Z=Y-\sigma^2$ is sub-exponential with parameter $(\nu,\alpha)=(2\sigma^2,4\sigma^2)$, so by Markov inequality
\begin{align*}
    P(Z\geq t)=&P(e^{\lambda Z}\geq e^{\lambda t})\\
    \leq& e^{-\lambda t}\cdot \mathbb{E}[e^{\lambda Z}]\\
    \leq&\operatorname{exp}\left(-\lambda t+2\lambda^2\sigma^4\right)
\end{align*}
when $t>\sigma^2=\frac{\nu^2}{\alpha}$, the function $-\lambda t+2\lambda^2\sigma^4$ monotonically decreasing, hence the tightest bound is reached at $\lambda = \alpha^{-1}=\frac{1}{4\sigma^2}$, and this bound becomes
\begin{align*}
    P(Z\geq t)\leq& \operatorname{exp}\left(-\frac{t}{\alpha}+\frac{1}{2\alpha}\frac{\nu^2}{\alpha}\right)\\
    \leq&\operatorname{exp}\left(-\frac{t}{\alpha}+\frac{t}{2\alpha}\right)\\
    =&\operatorname{exp}\left(-\frac{t}{2\alpha}\right)
\end{align*}
And thus complete the proof.
\qed
\\

\textbf{Technical Lemma B.6} \textit{Let $\Mcal{O}_{d_1,d_2}=\{V\in\mathbb{R}^{d_1\times d_2}|V^{\top}V=I\},(d_1\geq d_2)$, and $\epsilon\in(0,1)$. Then there exists a subset $\Mcal{N}\subset \Mcal{O}_{d_1,d_2}$ that is an $\epsilon-$net of $\Mcal{O}_{d_1,d_2}$ in Frobenius norm such that $|\Mcal{N}|\leq \left(\frac{6\sqrt{d_2}}{\epsilon}\right)^{d_1d_2}$, i.e. for any $V\in \Mcal{O}_{d_1,d_2}$, there exists a $V'\in \Mcal{N}$ such that $\|V-V'\|_F\leq \epsilon$.}

\textit{Proof.} For any $V\in \Mcal{O}_{d_1, d_2}$, each column of $V$ has unit $\ell_2$ norm. It is well known that there exists an $\frac{\epsilon}{2\sqrt{d_2}}$-net (in $\ell_2$ norm) of the unit sphere in $\mathbb{R}^{d_1}$ with size $\left(\frac{6\sqrt{d_2}}{\epsilon}\right)^{d_1}$. Using this net to cover all the columns, we obtain a set $\Mcal{N}' \subset \mathbb{R}^{d_1\times d_2}$ that is an $\frac{\epsilon}{2}$-net of $\Mcal{O}_{d_1, d_2}$ in Frobenius norm and $|\Mcal{N}'| \leq \left( \frac{6\sqrt{d_2}}{\epsilon}\right)^{d_1 d_2}$. 

After that, we transform $\Mcal{N}'$ into an $\epsilon$-net $\Mcal{N}$ that is a subset of $\Mcal{O}_{d_1,d_2}$. To achieve this, we just need to project each point in $\Mcal{N}'$ onto $\Mcal{O}_{d_1,d_2}$. Namely, for each $\bar{V}\in\Mcal{N}'$, let $\Mcal{P}(\bar{V})$ be its closest point in $\Mcal{O}_{d_1, d_2}$ in Frobenius norm; then define $\Mcal{N}=\left\{ \Mcal{P}(\bar{V}): \bar{V}\in \Mcal{N}' \right\}$. Then we have $|\Mcal{N}|\leq |\Mcal{N}'| \leq \left( \frac{6\sqrt{d_2}}{\epsilon} \right)^{d_1 d_2}$ and $\Mcal{N}$ is an $\epsilon$-net of $\Mcal{O}_{d_1, d_2}$, because for any $V \in \Mcal{O}_{d_1,d_2}$, there exists $\bar{V}\in \Mcal{N}'$ such $\|V-\bar{V}\|_F \leq \frac{\epsilon}{2}$, which implies $\Mcal{P}(\bar{V}) \in \Mcal{N}$ and $\|V-\mathcal{P}(\bar{V})\|_{F} \leq\|V-\bar{V}\|_{F}+\|\bar{V}-\mathcal{P}(\bar{V})\|_{F} \leq\|V-\bar{V}\|_{F}+\|\bar{V}-V\|_{F} \leq \epsilon$.
\qed 
\section{Appendix B. General Representation Class Analysis}
\label{general phi}
\subsection{Basic lemma and definitions}
We use following definition to characterize the expected difference between two general representation function.

\textbf{Definition C.0} (divergence between two representations) \textit{Given a distribution $q$ over $\mathbb{R}^D$ and two representation $\phi,\phi'\in \Phi$, the divergence between $\phi$ and $\phi'$ with respect to $q$ is defined as}
\begin{align*}
D_{q}\left(\phi, \phi^{\prime}\right)=\Sigma_{q}\left(\phi^{\prime}, \phi^{\prime}\right)-\Sigma_{q}\left(\phi^{\prime}, \phi\right)\left(\Sigma_{q}(\phi, \phi)\right)^{\dagger} \Sigma_{q}\left(\phi, \phi^{\prime}\right) \in \mathbb{R}^{d \times d}.
\end{align*}
Recall the definition of $\Sigma_{q}(\phi,\phi^{\prime})$ is
\begin{align*}
    \Sigma_{q}\left(\phi, \phi^{\prime}\right)=\mathbb{E}_{x \sim q}\left[\phi(x) \phi^{\prime}(x)^{\top}\right] \in \mathbb{R}^{d \times d}
\end{align*}
It is easy to verify that $D_q(\phi,\phi^{\prime}) \succeq 0$ and $D_q(\phi, \phi)=0$ for any $\phi,\phi^{\prime}$ and $q$. The next lemma reveals its connection to (symmetric) covariance.

\textbf{Lemma C.1} \textit{Suppose that that two representations $\phi,\phi^{\prime}\in \Phi$ and two distributions over $\mathbb{R}^D$ satisfy $\Lambda_{q}\left(\phi, \phi^{\prime}\right) \succeq \alpha \cdot \Lambda_{q^{\prime}}\left(\phi, \phi^{\prime}\right)$ for some $\alpha>0$. Then it must hold that }
\begin{align*}
    D_{q}\left(\phi, \phi^{\prime}\right) \succeq \alpha \cdot D_{q^{\prime}}\left(\phi, \phi^{\prime}\right)
\end{align*}
\textit{Proof.} Fix any $\bsyb{v}\in \mathbb{R}^D$, we will prove that $ \bsyb{v}^{\top} D_{q}\left(\phi, \phi^{\prime}\right) \bsyb{v} \geq \alpha \cdot \bsyb{v}^{\top} D_{q^{\prime}}\left(\phi, \phi^{\prime}\right) \bsyb{v}$, which proves the lemma.

We define a quadratic function $f:\mathbb{R}^D \mapsto \mathbb{R}$ as $f(\boldsymbol{w})=\left[\boldsymbol{w}^{\top},-\boldsymbol{v}^{\top}\right] \Lambda_{q}\left(\phi, \phi^{\prime}\right)\left[\begin{array}{c}
\boldsymbol{w} \\
-\boldsymbol{v}
\end{array}\right]$. According to the definition of symmetric covariance we know
\begin{align*}
f(\boldsymbol{w}) &=\boldsymbol{w}^{\top} \Sigma_{q}(\phi, \phi) \boldsymbol{w}-2 \boldsymbol{w}^{\top} \Sigma_{q}\left(\phi, \phi^{\prime}\right) \boldsymbol{v}+\boldsymbol{v}^{\top} \Sigma_{q}\left(\phi^{\prime}, \phi^{\prime}\right) \boldsymbol{v} \\
&=\mathbb{E}_{\boldsymbol{x} \sim q}\left[\left(\boldsymbol{w}^{\top} \phi(\boldsymbol{x})-\boldsymbol{v}^{\top} \phi^{\prime}(\boldsymbol{x})\right)^{2}\right]
\end{align*}
Therefore we know $f(\bsyb{w})\geq 0$ for any $\bsyb{w} \in \mathbb{R}^d$. This means that $f(\cdot)$ has a global minimizer in $\mathbb{R}^d$. Notice that $f$ is convex, this minimizer $\bsyb{w}^*$ is exact the point satisfying $\nabla f(\boldsymbol{w})=2 \Sigma_{q}(\phi, \phi) \boldsymbol{w}-2 \Sigma_{q}\left(\phi, \phi^{\prime}\right) \boldsymbol{v}=\bsyb{0}$, which can be calculated as $\boldsymbol{w}^{*}=\left(\Sigma_{q}(\phi, \phi)\right)^{\dagger} \Sigma_{q}\left(\phi, \phi^{\prime}\right) \boldsymbol{v}$. Plugging this into definition of function $f$, we have
$$
\min _{\boldsymbol{w} \in \mathbb{R}^{d}} f(\boldsymbol{w})=f\left(\boldsymbol{w}^{*}\right)=\boldsymbol{v}^{\top} D_{q}\left(\phi, \phi^{\prime}\right) \boldsymbol{v}.
$$
Similarly, letting $
g(\boldsymbol{w})=\left[\boldsymbol{w}^{\top},-\boldsymbol{v}^{\top}\right] \Lambda_{q^{\prime}}\left(\phi, \phi^{\prime}\right)\left[\begin{array}{c}
\boldsymbol{w} \\
-\boldsymbol{v}
\end{array}\right]
$, we have
$$
\min _{\boldsymbol{w} \in \mathbb{R}^{k}} g(\boldsymbol{w})=\boldsymbol{v}^{\top} D_{q^{\prime}}\left(\phi, \phi^{\prime}\right) \boldsymbol{v}.
$$
From $\Lambda_{q}\left(\phi, \phi^{\prime}\right) \succeq \alpha \cdot \Lambda_{q^{\prime}}\left(\phi, \phi^{\prime}\right)$ we know $f(\bsyb{w})\geq \alpha g(\bsyb{w})$ for any $\bsyb{w} \in \mathbb{R}^d$. Recall that $\bsyb{w}^*\in \arg\min_{\bsyb{w}\in\mathbb{R}^{d}} f(\bsyb{w})$. We have
\begin{align*}
\alpha \boldsymbol{v}^{\top} D_{q^{\prime}}\left(\phi, \phi^{\prime}\right) \boldsymbol{v} &=\underset{\boldsymbol{w} \in \mathbb{R}^{d}}{\alpha \min } g(\boldsymbol{w}) \leq \alpha g\left(\boldsymbol{w}^{*}\right) \leq f\left(\boldsymbol{w}^{*}\right) \\
&=\min _{\boldsymbol{w} \in \mathbb{R}^{d}} f(\boldsymbol{w})=\boldsymbol{v}^{\top} D_{q}\left(\phi, \phi^{\prime}\right), \boldsymbol{v}
\end{align*}
and finish the proof.
\qed
\\

\textbf{Lemma C.2} (In-sample Error Guarantee)
\textit{Let $\hat{\phi}$ and $\hat{\bsyb{w}}_1,\hdots,\hat{\bsyb{w}}_T$ be the solution obtained in algorithm 1. Denote the optimal ground truth representation function as $\phi^*(\cdot)$ and the optimal weight vector for task $t$ as $\bsyb{w}^*_t$. Then with probability at least $1-\delta$ we have}
\begin{align*}
    \sum_{t=1}^T &\left\| \hat{\phi}(X_t) \hat{\bsyb{w}}_t - \phi^*(X_t) \bsyb{w}^*_t \right\|^2 \lesssim (\sigma+1)^2\left(\Mcal{G}(\Mcal{F}_{\Mcal{X}}(\Phi))^2+\log\frac{1}{\delta}\right)
\end{align*}
\textit{Proof.} By the optimality of $\hat{\phi}$ and $\hat{\bsyb{w}}_1,\hdots,\hat{\bsyb{w}}_T$, we know
\begin{align*}
    \sum_{t=1}^T \left\|\hat{\phi}(X_t)\hat{\bsyb{w}}_t-\bsyb{y}_t\right\|^2\leq  \sum_{t=1}^T \left\|\phi^*(X_t)\bsyb{w}^*_t-\bsyb{y}_t\right\|^2
\end{align*}
Plug in $\bsyb{y}_t=\phi^*(X_t)\bsyb{w}^*_t+z_{t}$ where $z_{t}$ is a $5(\sigma+1)^2$ sub-Gaussian random variable independent of $X_t$, we get
\begin{align*}
     \sum_{t=1}^T \left\|\hat{\phi}(X_t)\hat{\bsyb{w}}_t-z_t-\phi^*(X_t)\bsyb{w}^*_t\right\|^2 \leq \sum_{t=1}^T \|z_t\|^2
\end{align*}
which gives 
\begin{align*}
     & \sum_{t=1}^T \left\|\hat{\phi}(X_t)\hat{\bsyb{w}}_t-\phi^*(X_t)\bsyb{w}^*_t\right\|^2\\
     \leq & 2\sum_{t=1}^T \langle z_t, \hat{\phi}(X_t)\hat{\bsyb{w}}_t-\phi^*(X_t)\bsyb{w}^*_t\rangle
\end{align*}
Denote $Z=[z_1,z_2,\hdots,z_T]\in \mathbb{R}^{N\times T}$ and $A=[a_1,a_2,\hdots,a_T]$ where each $a_t=\hat{\phi}(X_t)\hat{\bsyb{w}}_t-\phi^*(X_t)\bsyb{w}^*_t$. The above inequality can be compactly represented as $\|A\|_F^2\leq 2\langle Z,A \rangle$. Notice that $\frac{A}{\|A\|_F}\in\Mcal{F}_{\Mcal{X}}(\Phi)$. So we can bound $\|A\|_F$ as
\begin{align}
    \|A\|_F\leq 2\left\langle Z,\frac{A}{\|A\|_F}\right\rangle \leq 2\sup_{\bar{A}\in \Mcal{F}_{\Mcal{X}}(\Phi)} \langle Z,\bar{A}\rangle
\end{align}
By definition, denote the standard variance of $z_t$ as $\sigma'\leq \sqrt{5}(\sigma+1)$, we have 
$$\mathbb{E}_{Z}\left[\sup_{\bar{A}\in \Mcal{F}_{\Mcal{X}}(\Phi)} \langle  Z/\sigma',\bar{A}\rangle \right]=\Mcal{G}(\Mcal{F}_{\Mcal{X}}(\Phi))$$
Hence we know the expectation of right handside of (9) is no larger than $\sqrt{5}(\sigma+1)\Mcal{G}(\Mcal{F}_{\Mcal{X}}(\Phi))$. By the property that $Z\mapsto \sup_{\bar{A}\in \Mcal{F}_{\Mcal{X}}(\Phi)} \langle Z,\bar{A}\rangle$ is 1-Lipschitz in Frobenius norm, wo know according to standard Gaussian concentration inequality we know
\begin{align*}
    \sup_{\bar{A}\in \Mcal{F}_{\Mcal{X}}(\Phi)} \langle Z/\sigma',\bar{A}\rangle& \leq \mathbb{E}\left[\sup_{\bar{A}\in \Mcal{F}_{\Mcal{X}}(\Phi)} \langle Z/\sigma',\bar{A}\rangle\right]+\sqrt{\log\frac{1}{\delta}}\\
    &= \Mcal{G}(\Mcal{F}_{\Mcal{X}}(\Phi))+\sqrt{\log\frac{1}{\delta}}
\end{align*}
Then plug it into (9) and take square to complete the proof.
\qed
\\

\textbf{Lemma C.3} \textit{Under the setting of theorem 2, with probability at least $1-\delta$ we have}
\begin{align*}
    \frac{1}{n}\left\|P_{\hat{\phi}_h \left(X_{T+1}\right)}^{\perp} \phi^* \left(X_{T+1}\right)\right\|_{F}^{2} \lesssim \frac{(\sigma+1)^{2}\left(\mathcal{G}\left(\mathcal{F}_{\mathcal{X}}(\Phi)\right)^{2}+\log \frac{1}{\delta}\right)}{N \sigma_{k}^{2}\left(\hat{W}^{*}_h \right)}
\end{align*}
\textit{Proof.} 
Let $\hat{p}_t$ be the empirical distribution over samples in $X_t(t\in[T+1])$. According to assumption $7.1$ and settings in theorem 2, we know the concentration inequality for covariance holds with probability at least $1-\delta$
\begin{align*}
&0.9 \Lambda_{p}\left(\phi, \phi^{\prime}\right) \preceq \Lambda_{\hat{p}_{t}}\left(\phi, \phi^{\prime}\right) \preceq 1.1 \Lambda_{p}\left(\phi, \phi^{\prime}\right), \quad \forall \phi, \phi^{\prime} \in \Phi, \forall t \in[T] \\
&0.9 \Lambda_{p}\left(\hat{\phi}_h, \phi^* \right) \preceq \Lambda_{\hat{p}_{T+1}}\left(\hat{\phi}_h, \phi^*\right) \preceq 1.1 \Lambda_{p}\left(\hat{\phi}, \phi^*\right).
\end{align*}
Notice that $\hat{\phi}_h$ and $\phi^*$ are independent of the samples from samples in target task, so $n \geq N_{\operatorname{point}}(\Phi,p,\delta/3)$ is sufficient for the second inequality above to holds with high probability. Using lemma C.1 we know 
$$\begin{array}{l}
0.9 D_{p}\left(\phi, \phi^{\prime}\right) \preceq D_{\hat{p}_{t}}\left(\phi, \phi^{\prime}\right) \preceq 1.1 D_{p}\left(\phi, \phi^{\prime}\right), \quad \forall \phi, \phi^{\prime} \in \Phi, \forall t \in[T] \\
0.9 D_{p}\left(\hat{\phi}_h, \phi^*\right) \preceq D_{\hat{p}_{T+1}}\left(\hat{\phi}_h, \phi^*\right) \preceq 1.1 D_{p}\left(\hat{\phi}_h, \phi^*\right).
\end{array}$$
By the optimality of $\hat{\phi}_h$ and $\hat{\bsyb{w}}_{t,h}$ for $t\in[T]$ of optimization, we know $\hat{\phi}_h \left(X_{t}\right) \hat{\boldsymbol{w}}_{t}=P_{\hat{\phi}_h \left(X_{t}\right)} \boldsymbol{y}_{t}=P_{\hat{\phi}_h \left(X_{t}\right)}\left(\phi^* \left(X_{t}\right) \boldsymbol{w}_{t,h}^{*}+ z_t\right)$

Using lemma C.2, we have following chain of inequalities
\begin{align*}
& (\sigma+1)^{2}\left(\mathcal{G}\left(\mathcal{F}_{\mathcal{X}}(\Phi)\right)^{2}+\log \frac{H}{\delta}\right) \\
\gtrsim & \sum_{t=1}^{T}\left\|\hat{\phi}_h\left(X_{t}\right) \hat{\boldsymbol{w}}_{t}-\phi^*\left(X_{t}\right) \hat{\boldsymbol{w}}_{t,h}^{*}\right\|^{2} \\
=& \sum_{t=1}^{T}\left\|P_{\hat{\phi}_h\left(X_{t}\right)}\left(\phi^*\left(X_{t}\right) \hat{\boldsymbol{w}}_{t,h}^{*}+\boldsymbol{z}_{t}\right)-\phi^*\left(X_{t}\right) \hat{\boldsymbol{w}}_{t,h}^{*}\right\|^{2} \\
=& \sum_{t=1}^{T}\left\|-P_{\hat{\phi}_h\left(X_{t}\right)}^{\perp} \phi^*\left(X_{t}\right) \hat{\boldsymbol{w}}_{t,h}^{*}+P_{\hat{\phi}_h\left(X_{t}\right)} \boldsymbol{z}_{t}\right\|^{2} \\
=& \sum_{t=1}^{T}\left(\left\|P_{\hat{\phi}_h\left(X_{t}\right)}^{\perp} \phi^*\left(X_{t}\right) \hat{\boldsymbol{w}}_{t,h}^{*}\right\|^{2}+\left\|P_{\hat{\phi}_h\left(X_{t}\right)} \boldsymbol{z}_{t}\right\|^{2}\right)\\
\geq& \sum_{t=1}^{T}\left\|P_{\hat{\phi}_h\left(X_{t}\right)}^{\perp} \phi^*\left(X_{t}\right) \hat{\boldsymbol{w}}_{t,h}^{*}\right\|^{2} \\
\end{align*}
By definition of $P_{\hat{\phi}_h(X_t)}$, we know that

\begin{align*}
& \sum_{t=1}^{T}\left\|P_{\hat{\phi}_h\left(X_{t}\right)}^{\perp} \phi^*\left(X_{t}\right) \hat{\boldsymbol{w}}_{t,h}^{*}\right\|^{2} \\
=&\sum_{t=1}^{T}\left(\hat{\boldsymbol{w}}_{t,h}^{*}\right)^{\top} \phi^*\left(X_{t}\right)^{\top}\left(I-\hat{\phi}_h\left(X_{t}\right)\left(\hat{\phi}_h\left(X_{t}\right)^{\top} \hat{\phi}_h\left(X_{t}\right)\right)^{\dagger} \hat{\phi}_h\left(X_{t}\right)^{\top}\right) \phi^*\left(X_{t}\right)\hat{\boldsymbol{w}}_{t,h}^{*} \\
=& N \sum_{t=1}^{T}\left(\hat{\boldsymbol{w}}_{t,h}^{*}\right)^{\top} D_{\hat{p}_{t}}\left(\hat{\phi}_h, \phi^*\right) \hat{\boldsymbol{w}}_{t,h}^{*} \\
\geq& 0.9 N \sum_{t=1}^{T}\left(\hat{\boldsymbol{w}}_{t,h}^{*}\right)^{\top} D_{p}\left(\hat{\phi}_h, \phi^*\right) \hat{\boldsymbol{w}}_{t,h}^{*} \\
=& 0.9 N\left\|\left(D_{p}\left(\hat{\phi}_h, \phi^*\right)\right)^{1 / 2} \hat{W}_h^*\right\|_{F}^{2} \\
\geq& 0.9 N\left\|\left(D_{p}\left(\hat{\phi}_h, \phi^*\right)\right)^{1 / 2}\right\|_{F}^{2} \sigma_{k}^{2}\left(\hat{W}_h^*\right) \\
=& 0.9 N \operatorname{Tr}\left[D_{p}\left(\hat{\phi}_h, \phi^*\right)\right] \sigma_{k}^{2}\left(\hat{W}_h^*\right) \\
\geq& \frac{0.9 N}{1.1} \operatorname{Tr}\left[D_{\hat{p}_{T+1}}\left(\hat{\phi}_h, \phi^*\right)\right] \sigma_{k}^{2}\left(\hat{W}_h^*\right) \\
=&\frac{0.9 N}{1.1 n}\left\|P_{\hat{\phi}_h\left(X_{T+1}\right)}^{\perp} \phi^*\left(X_{T+1}\right)\right\|_{F}^{2} \sigma_{k}^{2}\left(\hat{W}_h^*\right).
\end{align*}
Completing the proof.
\qed
\\

\textbf{Lemma C.4} (Analogous to key lemma in linear cases) \textit{Denote $\kappa=\lambda_{\min}^{-1}(\Sigma_q(\hat{\phi},\hat{\phi}))$. If we have expected error bound within distribution $q$ is bounded by}
\begin{align*}
    \frac{1}{2} \mathbb{E}_{x\sim q} \left[\hat{\phi}(x)^{\top} \hat{\bsyb{w}} - \phi^*(x)^{\top}\bsyb{w}^*\right]^2 \leq \zeta_1
\end{align*}
and
\begin{align*}
    \operatorname{Tr}\left[D_q(\hat{\phi}, \phi^*) \right]\leq \zeta_2
\end{align*}
Then we know for any $\|x\|\leq 1$,
\begin{align*}
    \left| \hat{\phi}(x)^{\top} \hat{\bsyb{w}} - \phi^*(x)^{\top} \bsyb{w}^* \right| \lesssim \sqrt{\kappa \zeta_1} + \|\hat{\bsyb{w}}^*\| \zeta_2
\end{align*}

\textit{proof.} Denote $P_{\hat{\phi}}=\Sigma_{q}(\hat{\phi},\hat{\phi}) (\Sigma_{q}(\hat{\phi},\hat{\phi}))^{\dagger}$. For any $\|x\|\leq 1$, we can decompose $\phi^*(x)$ into $\phi^*(x)=P_{\hat{\phi}}\phi^*(x)+P_{\hat{\phi}}^{\perp} \phi^*(x)$. Then we have
\begin{align*}
     \left| \hat{\phi}(x)^{\top} \hat{\bsyb{w}} - \phi^*(x)^{\top} \bsyb{w}^* \right| &\leq  \left|\phi^*(x)^{\top} P_{\hat{\phi}}  \bsyb{w}^* - \hat{\phi}(x)^{\top} \hat{\bsyb{w}} \right| + \left| \phi^*(x)^{\top}  P_{\hat{\phi}}^{\perp} \bsyb{w}^* \right| \tag{$\star$}
\end{align*}
According to assumption 6.3, we know the second term is bounded by
\begin{align*}
     &\left\| \phi^*(x)^{\top}  P_{\hat{\phi}}^{\perp} \bsyb{w}^* \right\|^2 \leq \left\| P_{\hat{\phi}}^{\perp} \phi^*(x) \right\|^2 \cdot \left\| \hat{\bsyb{w}}^* \right\|^2 \leq \zeta_2^2 \|\hat{\bsyb{w}}^*\|^2
\end{align*}
Then we give a bound for the first term of $(\star)$. First derive the expected error within the space of $\operatorname{span}(\hat{\phi})$ as
\begin{align*}
    & \mathbb{E}_{x\sim q} \left|\phi^*(x)^{\top}  P_{\hat{\phi}}  \bsyb{w}^* - \hat{\phi}(x)^{\top} \hat{\bsyb{w}} \right|^2 \\
    =& \mathbb{E}_{x\sim q} \left|\phi^*(x)^{\top} \bsyb{w}^* - \phi^*(x)^{\top}  P_{\hat{\phi}}^{\perp} \bsyb{w}^* - \hat{\phi}(x)^{\top} \hat{\bsyb{w}} \right|^2 \\
    \leq & 2 \times \left(\left|\phi^*(x)^{\top} \bsyb{w}^* - \hat{\phi}(x)^{\top} \hat{\bsyb{w}} \right|^2 + \left| \phi^*(x)^{\top}  P_{\hat{\phi}}^{\perp}  \bsyb{w}^* \right|^2 \right)\\
    \leq & 2 (\zeta_1 + \left\|  \bsyb{w}^* \right\|^2 \zeta_2^2 )
\end{align*}
Applying assumption 6.4, we know there exist an invertible matrix $P$, such that for any $\|x\|\leq 1$ we have
\begin{align*}
   \|P\hat{\phi}(x)-\phi^*(x)\|^2 =& o(\zeta_1 ) / \| \bsyb{w}^* \|^2
\end{align*}
Therefore, we have following bound
\begin{align*}
    & \mathbb{E}_{x\sim q} \left|\phi^*(x)^{\top} P_{\hat{\phi}} \bsyb{w}^* - \hat{\phi}(x)^{\top} \hat{\bsyb{w}} \right|^2 \\
  = & \mathbb{E}_{x\sim q} \left| \phi^*(x)^{\top} P_{\hat{\phi}} \bsyb{w}^* - \hat{\phi}(x)^{\top} P^{\top} P_{\hat{\phi}}  \bsyb{w}^* + \hat{\phi}(x)^{\top} P^{\top} P_{\hat{\phi}}  \bsyb{w}^* - \hat{\phi}(x)^{\top} \hat{\bsyb{w}} \right|^2 \\
  = &  \mathbb{E}_{x\sim q} \left| (\phi^*(x) - P\hat{\phi}(x))^{\top} P_{\hat{\phi}} \bsyb{w}^* + \hat{\phi}(x)^{\top} (P^{\top} P_{\hat{\phi}}  \bsyb{w}^* - \hat{\bsyb{w}}) \right|^2  \\
  \geq& \mathbb{E}_{x\sim q} \left| \hat{\phi}(x)^{\top} (P^{\top} P_{\hat{\phi}}  \bsyb{w}^* - \hat{\bsyb{w}}) \right|^2 - o(\zeta_1)
\end{align*}
Denote $\Theta=P^{\top} P_{\hat{\phi}} \bsyb{w}^* - \hat{w}$ and take the spectral decomposition of matrix $\Sigma_q(\hat{\phi},\hat{\phi})=\mathbb{E}_{x \sim q}[\hat{\phi}(x)\hat{\phi}(x)^{\top}] = \sum_{j=1}^d \lambda_j \bsyb{v}_j \bsyb{v}_j^{\top}$ where $\bsyb{v}_j$ are othornormal vectors. Then we have
\begin{align*}
    & 2 (\zeta_1 + \left\|  \bsyb{w}^* \right\|^2 \zeta_2^2 )\\
    \geq & \mathbb{E}_{x\sim q} \left|\phi^*(x)^{\top}  P_{\hat{\phi}}  \bsyb{w}^* - \hat{\phi}(x)^{\top} \hat{\bsyb{w}} \right|^2 \\
    \geq& \mathbb{E}_{x\sim q} \left| \hat{\phi}(x)^{\top} (P^{\top} P_{\hat{\phi}}  \bsyb{w}^* - \hat{\bsyb{w}}) \right|^2 - o(\zeta_1) \\
    =& \Theta^{\top} \Sigma_q(\hat{\phi}, \hat{\phi}) \Theta - o(\zeta_1) \\
    =& \sum_{i=1}^{d} \lambda_{i}\left\|\boldsymbol{v}_{i}^{\top} \Theta\right\|^{2} - o(\zeta_1) \\
    \geq & \lambda_{\min}(\Sigma_q(\hat{\phi},\hat{\phi})) \sum_{i=1}^{d} \left\|\boldsymbol{v}_{i}^{\top} \Theta\right\|^{2} - o(\zeta_1)
\end{align*}
The above inequality can be summarized as $\sum_{i=1}^{d} \left\|\boldsymbol{v}_{i}^{\top} \Theta\right\|^{2} \leq 2 \kappa (\zeta_1 + \left\|  \bsyb{w}^* \right\|^2 \zeta_2^2 )$. So for arbitrary $x$, we write $\hat{\phi}(x)=\sum_{i=1}^d c_i \bsyb{v}_i$
\begin{align*}
&\left|\phi^*(x)^{\top} P_{\hat{\phi}} \bsyb{w}^* - \hat{\phi}(x)^{\top} \hat{\bsyb{w}} \right|^2 \\
  =& \left| \phi^*(x)^{\top} P_{\hat{\phi}} \bsyb{w}^* - \hat{\phi}(x)^{\top} P^{\top} P_{\hat{\phi}}  \bsyb{w}^* + \hat{\phi}(x)^{\top} P^{\top} P_{\hat{\phi}}  \bsyb{w}^* - \hat{\phi}(x)^{\top} \hat{\bsyb{w}} \right|^2 \\
 =&o(\zeta_1) + \Theta^{\top} \hat{\phi}(x) \hat{\phi}(x)^{\top} \Theta \\
=&o(\zeta_1)+\sum_{i=1}^{d} c_{i}^{2} \Theta^{\top} \boldsymbol{v}_{i} \boldsymbol{v}_{i}^{\top} \Theta \\
=&o(\zeta_1)+\sum_{i=1}^{d} c_{i}^{2}\left\|\Theta^{\top} \boldsymbol{v}_{i}\right\|^{2} \\
\leq& o(\zeta_1)+ C^2 \sum_{i=1}^{d} \left\|\boldsymbol{v}_{i}^{\top} \Theta\right\|^{2} \tag{$C=\|\hat{\phi}(x)\|$ is constant bounded} \\
\lesssim & \kappa (\zeta_1 + \left\|  \bsyb{w}^* \right\|^2 \zeta_2^2 )
\end{align*}
Back to ($\star$) and plug in the bound for both terms, we get
\begin{align*}
     \left| \hat{\phi}(x)^{\top} \hat{\bsyb{w}} - \phi^*(x)^{\top} \bsyb{w}^* \right| &\leq  \left|\phi^*(x)^{\top} P_{\hat{\phi}}  \bsyb{w}^* - \hat{\phi}(x)^{\top} \hat{\bsyb{w}} \right| + \left| \phi^*(x)^{\top}  P_{\hat{\phi}}^{\perp} \bsyb{w}^* \right| \\
     \lesssim&  \sqrt{ \kappa (\zeta_1 + \left\|  \bsyb{w}^* \right\|^2 \zeta_2^2 )} + \left\| \bsyb{w}^* \right\| \zeta_2 \\
     \lesssim& \sqrt{\kappa \zeta_1} + \|\hat{\bsyb{w}}^*\| \zeta_2
\end{align*}
And complete the proof. 
\qed
\\

\subsection{Proof of Theorem 2}
For the sake of simplicity, we will not track the constant of probability $\delta$, and use $1-O(\delta)$ instead.

In this section we will prove main theorem. The overall structure is similar to linear representation. Namely, we just need to show at each level $h$, for any state-action pair $(s,a) \in \Mcal{S} \times \Mcal{A}$, $x=\xi(s,a)$
\begin{align}
    \left| \hat{\phi}_h(x) \hat{\bsyb{w}} - \phi^*(x) \hat{\bsyb{w}}^*_h \right| \lesssim \epsilon_h,
\end{align}
holds with probability at least $1-O(\delta/H)$ for some $\epsilon > 0$, then applying the same procedure as in appendix A for linear representation, we have
\begin{align*}
    \left| V^{\star}_h(s) - V^{\pi}_h(s) \right| \lesssim \sum_{k=h}^H \epsilon_k.
\end{align*}
Therefore we will focus on proving (14) of which detailed $\epsilon_h$ is as 
\begin{align*}
     \left| \hat{\phi}_h(x) \hat{\bsyb{w}} - \phi^*(x) \hat{\bsyb{w}}^*_h \right| \leq&  \sqrt{ \frac{\kappa (d+\log(H/\delta))}{n} + \kappa \frac{\Mcal{G}(\Mcal{F}_{\Mcal{X}}(\Phi))^2 d+\log(H/\delta)}{N T} } \\
     \lesssim& \sqrt{ \frac{\kappa (d+\log(H/\delta))}{n} + \frac{ \Mcal{G}(\Mcal{F}_{\Mcal{X}}(\Phi))^2 \kappa d}{N T} } 
\end{align*}

First, we establish the bound for excess risk $\operatorname{ER}_h(\hat{\phi}_h. \hat{\bsyb{w}}_{T+1,h})$ by applying Lemma C.3 
\begin{align*}
& \operatorname{ER}_h \left(\hat{\phi}_h, \hat{\boldsymbol{w}}_{T+1,h}\right) \\
=& \frac{1}{2} \mathbb{E}_{\boldsymbol{x} \sim p}\left[\left(\hat{\boldsymbol{w}}_{T+1,h}^{\top} \hat{\phi}_h(\boldsymbol{x})-\left(\hat{\boldsymbol{w}}_{T+1,h}^{*}\right)^{\top} \phi^*(\boldsymbol{x})\right)^{2}\right] \\
=& \frac{1}{2}\left[\begin{array}{c}
\hat{\boldsymbol{w}}_{T+1,h} \\
-\hat{\boldsymbol{w}}_{T+1,h}^{*}
\end{array}\right]^{\top} \Lambda_{p}\left(\hat{\phi}_h, \phi^*\right)\left[\begin{array}{c}
\hat{\boldsymbol{w}}_{T+1,h} \\
-\hat{\boldsymbol{w}}_{T+1,h}^{*}
\end{array}\right] \\
\lesssim &\left[\begin{array}{c}
\hat{\boldsymbol{w}}_{T+1,h} \\
-\hat{\boldsymbol{w}}_{T+1,h}^{*}
\end{array}\right]^{\top} \Lambda_{\hat{p}_{T+1,h}}\left(\hat{\phi}_h, \phi^*\right)\left[\begin{array}{c}
\hat{\boldsymbol{w}}_{T+1,h} \\
-\hat{\boldsymbol{w}}_{T+1,h}^{*}
\end{array}\right] \\
=& \frac{1}{n}\left\|\hat{\phi}_h\left(X_{T+1}\right) \hat{\boldsymbol{w}}_{T+1,h}-\phi^*\left(X_{T+1}\right) \hat{\boldsymbol{w}}_{T+1,h}^{*}\right\|^{2} \\
=& \frac{1}{n}\left\|-P_{\hat{\phi}_h\left(X_{T+1}\right)}^{\perp} \phi^*\left(X_{T+1}\right) \hat{\boldsymbol{w}}_{T+1,h}^{*}+P_{\hat{\phi}_h\left(X_{T+1}\right)} \boldsymbol{z}_{T+1}\right\|^{2} \\
=& \frac{1}{n}\left(\left\|P_{\hat{\phi}_h\left(X_{T+1}\right)}^{\perp} \phi^*\left(X_{T+1}\right) \hat{\boldsymbol{w}}_{T+1,h}^{*}\right\|^{2}+\left\|P_{\hat{\phi}_h\left(X_{T+1}\right)} \boldsymbol{z}_{T+1}\right\|^{2}\right) \\
\lesssim & \frac{1}{n}\left\|P_{\hat{\phi}_h\left(X_{T+1}\right)}^{\perp} \phi^*\left(X_{T+1}\right) \hat{\boldsymbol{w}}_{T+1,h}^{*}\right\|^{2}+\frac{(1+\sigma)^{2}\left(d+\log \frac{H}{\delta}\right)}{n} \\
\lesssim&  \frac{(\sigma+1)^{2}\left(\mathcal{G}\left(\mathcal{F}_{\mathcal{X}}(\Phi)\right)^{2}+\log \frac{H}{\delta}\right)  \|\hat{\bsyb{w}}_{T+1,h}^2\|^2}{N \sigma_{d}^{2}\left(\hat{W}^{*}_h\right)}+\frac{(\sigma+1)^{2}\left(d+\log \frac{H}{\delta}\right)}{n} \\
\lesssim& \frac{(\sigma+1)^{2}\left(\mathcal{G}\left(\mathcal{F}_{\mathcal{X}}(\Phi)\right)^{2}d+d\log \frac{H}{\delta}\right)}{N T}+\frac{(\sigma+1)^{2}\left(d+\log \frac{H}{\delta}\right)}{n}
\end{align*}
Also, another by-product of lemma C.3 is the upper bound for $D_{p} \left( \hat{\phi}_h,\phi^* \right)$ as
\begin{align*}
    \operatorname{Tr}\left[ D_{p}\left(\hat{\phi}_h, \phi^*\right) \right] 
    \lesssim  \frac{(\sigma+1)^{2}\left(\mathcal{G}\left(\mathcal{F}_{\mathcal{X}}(\Phi)\right)^{2}+\log \frac{H}{\delta}\right)}{N \sigma_{d}^{2}\left(\hat{W}^{*}_h\right)}
\end{align*}

Similar to the analysis in section A, to make sure that $\sigma_{d}^{2}\left(\hat{W}^{*}_h\right)=\Omega(T(H-h+1)^2/d)$, we just need to make sure the number of samples at next level $h+1$ to be big enough, for now we can assume that holds and add condition for the number of samples. Thus according to lemma C.4, we know for any $x$ such that $\|x\| \leq 1$, the uniform prediction error is bounded by
\begin{align*}
    &\left\| \hat{\phi}_h(x)^{\top} \hat{\bsyb{w}}_{T+1,h} - \phi^*(x)^{\top} \hat{\bsyb{w}}^*_{T+1,h} \right\|\\
    \lesssim & \|\hat{\bsyb{w}}_{T+1,h}^*\|\sqrt{\frac{(\sigma+1)^{2}d \left(\mathcal{G}\left(\mathcal{F}_{\mathcal{X}}(\Phi)\right)^{2}+\log \frac{H}{\delta}\right)}{N T (H-h+1)^2}} \\
    &+ \sqrt{\kappa \left( \frac{(\sigma+1)^{2}d\left(\mathcal{G}\left(\mathcal{F}_{\mathcal{X}}(\Phi)\right)^{2}+\log \frac{H}{\delta}\right)}{N T}+\frac{(\sigma+1)^{2}\left(d+\log \frac{H}{\delta}\right)}{n}\right) }\\
\end{align*}
Hence we get 
\begin{align*}
    &\left\| \hat{\phi}_h(x)^{\top} \hat{\bsyb{w}}_{T+1,h} - \phi^*(x)^{\top} \hat{\bsyb{w}}^*_{T+1,h} \right\|\\
    \lesssim & (\sigma+1) \sqrt{\kappa} \sqrt{\frac{\mathcal{G}\left(\mathcal{F}_{\mathcal{X}}(\Phi))^{2}d+\log \frac{H}{\delta}\right)}{N T} + \frac{ d+\log \frac{H}{\delta}}{n}}  \\
\end{align*}
Notice $T>d$, hence to make sure that the value function is accurate, we need each level's $N\gg \kappa \mathcal{G}\left(\mathcal{F}_{\mathcal{X}}(\Phi)\right)^{2}$ and $n \gg \kappa d$. Then analogous to the procedure in Appendix A, we get the sub-optimality for learned policy $\pi_h(s)=\arg\max_{a\in\Mcal{A}} \hat{\phi}_h(s,a)^{\top} \hat{\bsyb{w}}_{T+1,H}$ is bounded by
\begin{align*}
    \left| V^{\star}_h(s) - V^{\pi}_h(s) \right| \lesssim \sqrt{\kappa} (\sigma+1)(H-h+1)\sqrt{\frac{\mathcal{G}\left(\mathcal{F}_{\mathcal{X}}(\Phi)\right)^{2}d+\log \frac{H}{\delta}}{N T} +\frac{(\sigma+1)^{2}\left(d+\log \frac{H}{\delta}\right)}{n}}  
\end{align*}
If regarding whole tuple $(s,a,r(s,a),s')$ as one sample, then we need to substitute $N$ by $N/H^2$ and $n$ by $n/H^2$, which gives the final result as 
\begin{align*}
    \left| V^{\star}(s) - V^{\pi}(s) \right| \lesssim \sqrt{\kappa}(\sigma+1)H^2\sqrt{\frac{\mathcal{G}\left(\mathcal{F}_{\mathcal{X}}(\Phi)\right)^{2}d+\log \frac{H}{\delta}}{N T} +\frac{d+\log \frac{H}{\delta}}{n}}  
\end{align*}
and complete the proof.
\qed
\\

\subsection{Technical Lemmas}
\textbf{Technical Lemma D.1} (Equivalence of Gaussian width and sub-Gaussian width) \textit{Given a set $S \subset \mathbb{R}^d$ and a $\rho^2$ sub-Gaussian distribution $P$. Then we know the sub-Gaussian width defined by}
\begin{align*}
    \Mcal{G}_{P}(S):=\mathbb{E}_{z\sim P^d}\left[\sup_{x\in S}\left\langle x,z \right\rangle\right]
\end{align*}
is $\rho$-bounded by standard Gaussian width $\Mcal{G}(S)$, namely
$$\Mcal{G}_{P}(S) \leq c \rho \cdot \Mcal{G}(S)$$
where $c>0$ is a fixed constant independent of $d$ and $S$.

\textit{Proof.} Denote the distribution of a standard d-dimensional Gaussian variable's norm, which is the square root of a $\chi^2(d)$ distribution as $\ell_{\Mcal{N}}$ (similarly $\ell_{P}$ for the norm distribution of variable sampled from $P$). By using conditional expectation we know
\begin{align*}
    \Mcal{G}(S) =& \mathbb{E}_{\ell \sim \ell_{\Mcal{N}}}\left[ \mathbb{E}_{\bsyb{u} \sim U(\Mcal{S}^{d-1})}\left[ \sup_{x\in S}\langle \ell \cdot \bsyb{u}, x \rangle \right] \Big| \|z\|=\ell \right] \\ 
    =& \mathbb{E}_{\ell \sim \ell_{\Mcal{N}}}\left[\ell \cdot \mathbb{E}_{\bsyb{u} \sim U(\Mcal{S}^{d-1})} \left[\sup_{x\in S}\langle \bsyb{u}, x \rangle \right] \Big| \|z\|=\ell \right]
\end{align*}
Define a constant $R_S$ with respect to set $S$ as
\begin{align*}
    R_S :=\mathbb{E}_{\bsyb{u} \sim U(\Mcal{S}^{d-1})} \left[\sup_{x\in S}\langle \bsyb{u}, x \rangle \right]
\end{align*}
Then we know a simple equation for Gaussian width is
\begin{align*}
    \Mcal{G}(S) =& R_S\cdot \mathbb{E}_{z\sim \ell_{\Mcal{N}}}[z]\\
    =& R_S \int_{0}^{+\infty} \sqrt{x} \cdot \frac{1}{2^{\frac{d}{2}}\cdot  \Gamma\left(\frac{d}{2}\right)} x^{\frac{d}{2}-1} e^{-\frac{x}{2}} \mathrm{d}_x \\
    =& R_S\cdot \frac{\sqrt{2}\Gamma\left(\frac{d+1}{2}\right)}{\Gamma\left(\frac{d}{2}\right)} 
\end{align*}
Since $\lim_{x\to \infty} \frac{\Gamma\left(\frac{d+1}{2}\right)}{\Gamma\left(\frac{d}{2} \right)d^{1/2}}=1$, we know $\frac{\sqrt{2}\Gamma\left(\frac{d+1}{2}\right)}{\Gamma\left(\frac{d}{2}\right)}$ is constant lower bounded. Therefore $\Mcal{G}(S) \geq c\sqrt{d}\cdot R_S$ for some fixed constant $c>0$ that is independent of $S$ and $d$.

According to classic result for sub-Gaussian variable's norm, we know that
$$\mathbb{E}_{z\sim P^d} [\|z\|] \leq 4\rho \sqrt{d}$$
Thus we know

\begin{align*}
    \Mcal{G}_P(S) =& \mathbb{E}_{\ell \sim \ell_{P}}\left[ \mathbb{E}_{\bsyb{u} \sim U(\Mcal{S}^{d-1})}\left[ \sup_{x\in S}\langle \ell \cdot \bsyb{u}, x \rangle \right] \Big| \|z\|=\ell \right] \\ 
    =& \mathbb{E}_{\ell \sim \ell_{P}}\left[\ell \cdot \mathbb{E}_{\bsyb{u} \sim U(\Mcal{S}^{d-1})} \left[\sup_{x\in S}\langle \bsyb{u}, x \rangle \right] \Big| \|z\|=\ell \right]\\
    =& R_S \cdot \mathbb{E}_{z \sim \ell_{P}}[z] \\
    \leq& 4 R_S\cdot \rho \sqrt{d} \\
    \leq& \frac{4}{c}\cdot\rho \Mcal{G}(S).
\end{align*}
Hence the claim of our lemma holds. 
\qed
\end{document}